\documentclass[10pt,twocolumn,letterpaper]{article}

\usepackage[pagenumbers]{cvpr}  %

\pdfinfoomitdate=1 \pdftrailerid{} \pdfsuppressptexinfo=-1

\usepackage{amsmath,amssymb,amsthm}
\usepackage{booktabs}
\usepackage{algorithm}
\usepackage{algpseudocode}
\usepackage{tikz}
\usetikzlibrary{positioning,arrows.meta,fit,calc}
\usepackage{xspace}
\newcommand{\mba}{MBA\xspace}
\newtheorem{proposition}{Proposition}

\newcommand{\Pb}{\mathbb{P}}
\newcommand{\E}{\mathbb{E}}
\newcommand{\N}{\mathcal{N}}
\newcommand{\one}{\mathbf{1}}
\definecolor{blueG}{HTML}{2471A3}
\definecolor{orangeP}{HTML}{E67E22}
\definecolor{greenO}{HTML}{27AE60}
\usetikzlibrary{backgrounds}

\definecolor{cvprblue}{rgb}{0.21,0.49,0.74}
\usepackage{array}
\usepackage[breaklinks,colorlinks,allcolors=cvprblue]{hyperref}

\title{Motion Belief Action: From a Visual Scene Map to Physical Decisions}
\author{Dibyendu Ghosh\\}

\begin{document}
\maketitle
\begin{abstract}
A robot that works among people must judge what it cannot see: whether an object is still where it was last seen, and whether a pull on a
door or drawer is worth its risk. Maps that track object change hold nothing about force, maps of physical properties
answer a risk question at a fixed risk level, and vision-language-action models keep no explicit estimate of hidden
physics. We propose Motion Belief Action (MBA), a decision layer on the scene map. From how often people move an object
and how much force it needs, how much load it holds and whether it is blocked, MBA chooses under a force limit to look
again, nudge, open, brace or decline, with its risk set by the task's costs. On real activity in 9 kitchens, it predicts
whether an object is still in place better than FreMEn and Perpetua, and decides when to look again at lower cost than
these and the persistence filter, because about a quarter of placements end in a pick-up within 30\,s. On real pulls
of doors and drawers in 17 locations, made by a person with a force-sensing gripper, the map's force estimates decide
better than a part-type average. Setting the risk from the costs also beats a published risk rule at its fixed level. In simulation, with the force limit at the training parts' median force, MBA raises
SmolVLA's and Octo's scores by declining pulls; most declined door pulls would fail, but on drawers more would have
opened, and a part-type average gets about half of this gain. In both, a published risk rule set to the risk these
costs imply is MBA's rule for a single pull and decides as well in the main settings. A link between opening and holding force, a probing
nudge and an action prior for camera tracking do not help in the main settings, and over the real-pull sweep the nudge costs a little.
\end{abstract}

\section{Introduction}
\label{sec:intro}

A robot that works among people must act on what it cannot see. People move things: a mug put down a moment ago
may be picked up again within a minute. Doors and drawers hide their
physics. Before pulling a drawer, the robot does not know how hard it will resist. Before leaning on an open door, it
does not know whether the door will hold. A pull with too much force can damage the part, so the robot
works under a force limit. For each, it can look again, act or decline, and for a door or drawer first test it; the
right choice depends on what a mistake costs in the task. The evidence is in the robot's scene map, which records what people move
and what the robot felt when it touched a part.

Maps that track change estimate whether an object is still in place. The persistence filter~\cite{rosen2016persistence} estimates how
likely a mapped object is to still be there, and FreMEn~\cite{krajnik2017fremen} models objects that come and go
periodically. Perpetua~\cite{saavedra2025perpetua} mixes several such filters with different lifetimes but predicts from the most
likely one, and builds them into scene graphs~\cite{saavedra2026predictive}; PreSIST~\cite{adkins2026presist} sets an object's prior from a
vision-language model; HiPer~\cite{canh2026hiper} pools per-object rates, and an earlier model adds a second component~\cite{toris2017temporal}. Open-vocabulary maps label every object~\cite{gu2024conceptgraphs}, VLMM~\cite{ghosh2026vlmm}
records which objects people move, and semantic exploration revisits places whose objects may have
moved~\cite{bogenberger2026glasses}, and active perception re-inspects changed objects~\cite{qian2026active}. These maps tell the robot where to look, but they hold nothing about force, so
they cannot tell the arm how to act.

Methods that learn by touching estimate a part's physics~\cite{ewen2024feel}. The joint of a door or drawer, its type and axis, is estimated from
the robot's own motion~\cite{sturm2011probabilistic,martinmartin2014online,hausman2015active}.
PhysVLMM~\cite{ghosh2026physvlmm} keeps an estimate of each part's forces in the scene map, shares it between parts
that look alike, and tells whether an action will succeed.
Phys2Real~\cite{wang2026phys2real} and Law-to-Action~\cite{mabrok2026l2a} combine vision-language priors with
estimates made by touch. Test actions reveal whether an object moves or a drawer is locked~\cite{yuan2026wontknow,liu2026aquamend}, TANav~\cite{li2026tanav} nudges an obstacle and decides from costs whether to push it, a force-capped pull is refused when it would move the furniture~\cite{zhou2026trading}, and
MessyMem~\cite{banwasi2026messymem} remembers what the robot learned by doing. Law-to-Action comes closest: it
chooses to look again or test by expected value under hand-set costs and refuses at a fixed threshold~\cite{mabrok2026l2a}. In our method,
every choice, to look again, test, act or decline, follows from the task's costs, as in the classical reject rule~\cite{chow1970optimum}, with no other threshold set by hand.

Vision-language-action models~\cite{kim2024openvla,octo2024,shukor2025smolvla,black2025pi0} map an image and an
instruction directly to motion, and active visual reasoning~\cite{zhou2025physvlmavr} lets a model move to see what is
hidden, and calibrated abstention refuses actions likely to be unsafe~\cite{singh2026bokbo,zhang2026rcdh}. They hold no explicit estimate of a part's force, so they cannot judge whether a test is worth its cost, or
decline a pull from the part's own estimated force.

Tracking the camera while the robot acts is a smaller problem: methods for moving scenes remove pixels of classes that
usually move~\cite{bescos2018dynaslam,murartal2017orbslam2} or track rigid objects~\cite{long2021rigidfusion}, but a drawer the robot pulls moves with the wrist camera, so its pixels stay still.

\begin{figure*}[t]
\centering
\resizebox{\textwidth}{!}{%
\begin{tikzpicture}[font=\footnotesize, >={Latex[length=2mm]},
  blk/.style={draw=#1!55!black, rounded corners=3pt, fill=#1!8, minimum width=2.4cm, minimum height=2.55cm, inner sep=0pt},
  ttl/.style={font=\footnotesize\bfseries, align=center},
  sub/.style={font=\scriptsize, align=center},
  arr/.style={->, thick, draw=black!75},
  lab/.style={font=\scriptsize, text=black!70, align=center}]
\newcommand{\thumb}[4]{%
  \node[ttl] at ($(#1.north)+(0,-0.27)$) {#3};
  \node at ($(#1.center)+(0,0.03)$) {\includegraphics[height=1.12cm, width=2.15cm, keepaspectratio]{#2}};
  \node[sub] at ($(#1.south)+(0,0.32)$) {#4};}
\node[inner sep=0pt] (frame) at (1.3,0) {\includegraphics[height=1.95cm]{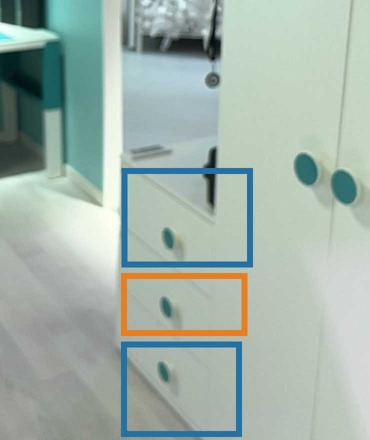}\,\includegraphics[height=1.95cm]{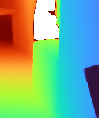}};
\node[ttl, above=2pt of frame] {RGB-D + poses};
\node[sub, below=2pt of frame] (sencap) {{\color{blueG}$\blacksquare$} look-alike drawers\ {\color{orangeP}$\blacksquare$} to open};
\node[blk=gray] (m1) at (4.3,0) {}; \node[ttl] at ($(m1.north)+(0,-0.27)$) {M1 VLMM};
\node[sub] at ($(m1.center)+(0,-0.05)$) {(frozen)\\[4pt] \footnotesize elements $e_i$:\\ \footnotesize label $\ell_i$,\\ \footnotesize category $h_i$,\\ \footnotesize motion class $y_i$};
\node[blk=blue] (m2) at (7.0,0) {}; \thumb{m2}{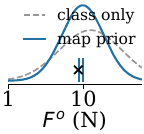}{M2 PhysVLMM}{force prior\\ from look-alikes};
\node[blk=greenO, fill=greenO!22, draw=greenO!65!black, line width=1.1pt, minimum width=2.6cm, minimum height=2.85cm] (m4) at (9.85,0) {};
\node[ttl, text=greenO!25!black] at ($(m4.north)+(0,-0.42)$) {M4 MBA belief\\[-1pt](ours)};
\node at ($(m4.center)+(0,-0.05)$) {\includegraphics[height=1.05cm]{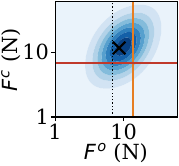}};
\node[sub] at ($(m4.south)+(0,0.36)$) {$b_{t,i}$: forces, blocked,\\ move rate};
\node[blk=green] (m5) at (12.7,0) {}; \thumb{m5}{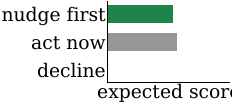}{M5 Planner (ours)}{exact values under\\ the force limit};
\node[blk=yellow, minimum width=1.75cm] (ans) at (15.05,0) {}; \node[ttl] at ($(ans.north)+(0,-0.27)$) {Answer};
\node[sub, font=\footnotesize] at ($(ans.center)+(0,-0.1)$) {look again /\\ nudge / open /\\ brace / decline\\[5pt] camera\\ rotation $\mathbf R_t$};
\node[inner sep=0pt] (ppl) at (1.3,-3.6) {\includegraphics[height=1.6cm]{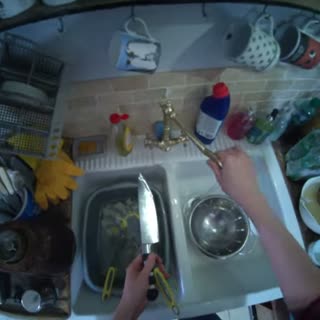}};
\node[ttl, above=2pt of ppl] {people move objects};
\node[sub, below=2pt of ppl] {HD-EPIC: pick-ups, put-downs};
\node[blk=blue] (m3) at (4.3,-3.6) {}; \thumb{m3}{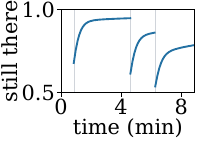}{M3 Still in place (ours)}{still in place? $P_i(\Delta)$\\ (one colander)};
\node[blk=orange] (act) at (10.4,-3.6) {}; \thumb{act}{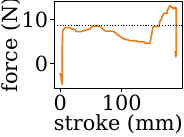}{M6 Action + F/T}{pull, nudge or brace;\\ touch outcome};
\node[blk=green] (trk) at (13.6,-3.6) {}; \thumb{trk}{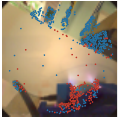}{M7 Tracking (ours)}{{\color{red!70!black}$\bullet$} moves with camera\\[-1pt] {\color{blueG}$\bullet$} world $\Rightarrow\mathbf R_t$};
\draw[arr] (frame.east) -- (m1.west); \draw[arr] (m1) -- (m2); \draw[arr] (m2) -- (m4); \draw[arr] (m4) -- (m5); \draw[arr] (m5) -- (ans);
\draw[arr, dashed, draw=black!55] (m1.south) -- node[lab, right]{label, category} (m1.south |- m3.north);
\draw[arr] (ppl.east) -- (m3.west);
\draw[arr] (m3.east) -- node[lab, above]{rest gaps, time at rest} ([xshift=-8mm]m4.south |- m3.east) -- ([xshift=-8mm]m4.south);
\draw[arr] ([xshift=-2mm]act.north) -- node[lab, left]{touch\\[-2pt] outcome} ([xshift=-2mm]act.north |- m4.south);
\draw[arr] (m5.south) -- ++(0,-0.6) -| node[lab, pos=0.25, above]{decision} ([xshift=6mm]act.north);
\draw[arr] (act) -- node[lab, above]{while\\[-2pt] holding} (trk);
\draw[arr] (trk.east) -| (ans.south);
\node[font=\small] (capa) at (8.0,-5.7) {(a)};
\draw[dashed, draw=black!30] (16.25,1.9) -- (16.25,-5.35);
\begin{scope}[xshift=17.95cm, yshift=0.0cm]
\node[inner sep=0] (b0) at (-0.15,0.35) {\includegraphics[width=1.75cm]{figs/arch/belief_prior.pdf}};
\node[inner sep=0] (b1) at (3.1,0.95) {\includegraphics[width=1.75cm]{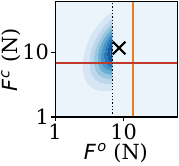}};
\node[inner sep=0] (b2) at (3.1,-1.85) {\includegraphics[width=1.75cm]{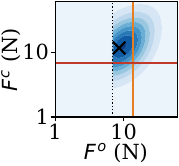}};
\node[sub] at ($(b0.north)+(0,0.12)$) {before};
\node[sub] at ($(b1.north)+(0,0.12)$) {moved ($p=0.29$)};
\node[sub] at ($(b2.north)+(0,0.12)$) {stayed ($p=0.71$)};
\draw[->, semithick] (b0.east) -- (b1.west); \draw[->, semithick] (b0.east) -- (b2.west);
\node[sub, align=left, anchor=north west] at ($(b0.south west)+(-0.05,-0.1)$) {$\bar p_o=0.70$\\ $p_b=0.73$\\ $V_N=0.375$\\ $V_O=0.394$\\ $\Rightarrow$ \textbf{open}};
\node[sub, align=left, anchor=west] at ($(b1.east)+(0.08,0)$) {$\bar p_o=1.00$\\ $p_b=0.62$\\ $\Rightarrow$ \textbf{open}};
\node[sub, align=left, anchor=west] at ($(b2.east)+(0.08,0)$) {$\bar p_o=0.57$\\ $p_b=0.80$\\ $\Rightarrow$ \textbf{open}\\ \textbf{+ brace}};
\node[sub, anchor=west] at (-1.2,-3.35) {\textcolor{orangeP}{---} $F_{\max}$ \quad \textcolor{red!80!black}{---} $F_L$ \quad $\cdots$ $f_n$ \quad $\times$ measured (held out)};
\node[draw=greenO!65!black, rounded corners=5pt, line width=1.1pt, fit={(-1.3,2.15) (5.5,-3.85)}, inner sep=3pt] (boxb) {};
\node[ttl, fill=white, inner sep=2pt] at (boxb.north) {How one nudge changes the decision};
\end{scope}
\node[font=\small] at (boxb.south |- capa) {(b)};
\end{tikzpicture}}
\caption{\textbf{Architecture of \mba} on real data. (a) Modules M1--M7, from the RGB-D input to the answer, on one
Hoi!~\cite{engelbracht2026hoi} location. The still-in-place model (M3) is shown on HD-EPIC~\cite{perrett2025hdepic}, as no
object in our data has both measured forces and people moving it. (b) The orange drawer's force belief before a possible nudge and after each
outcome ($p$: chance of the outcome), with the planner's quantities of \eqref{eq:vbrace}--\eqref{eq:vnudge}: $\bar p_o$,
probability that a pull opens it; $p_b$, probability that a brace holds after opening; $V_N$, $V_O$, values of
nudging first and acting now.}
\label{fig:architecture}
\end{figure*}

We present \emph{Motion Belief Action} (\mba), a decision layer on the scene map (\cref{fig:architecture}). For each
object, \mba keeps a belief, a probability distribution, over how often people move
it, the force needed to open it, the load it holds when the robot leans on it (a \emph{brace}), and whether it is
blocked. The force estimates start from PhysVLMM's~\cite{ghosh2026physvlmm}. \mba adds the rest: a still-in-place model, the blocked
state, updates from whether a touch moved or held the part, the planner and the camera tracker. Under the force limit, the planner chooses whether to look again, test the force with a
light push (a \emph{nudge}), open, brace or decline. It weighs each choice by its expected outcome under the task's costs, so the risk it takes follows the task. PhysVLMM's risk query instead takes the risk level as an input;
for a single pull, with that level set from the same costs, the two choose alike. We test each part on its own data. Our contributions are:
\begin{enumerate}
\item \textbf{\mba} (\cref{sec:method}). A decision layer in which every choice follows from the task's costs. On real pulls, the map's force estimates decide better than a part-type average
  (\cref{sec:r-pulls}); in simulation it raises the scores of two vision-language-action policies, mostly by declining
  pulls (\cref{sec:r-closed}).
\item \textbf{Still-in-place prediction} (\cref{sec:belief}). One quick pick-up shared by all objects, since people often pick an
  object up again soon after putting it down, and a slow rate per object, estimated together for objects with the same name. On real kitchen activity, its decisions to look again cost less than those of three published methods
  (\cref{sec:r-presence}).
\item \textbf{Evaluation} (\cref{sec:results}). Real pulls, real kitchen activity and simulation, against published
  methods run from their authors' code where released. It also reports what does not help: the link between opening
  force and held load, the nudge, and the tracker's prior on where the held part appears.
\end{enumerate}

\section{Methodology}
\label{sec:method}

\subsection{Problem formulation}
\label{sec:formulation}

\paragraph{Scene map.}
\mba runs on a Vision-Language-Motion Map (VLMM)~\cite{ghosh2026vlmm}, which gives each object its name and records
whether and when people move it. Each map element $e_i$ has a position $\mathbf{p}_i$, an open-vocabulary label
$\ell_i$, a category $h_i$ and a motion class $y_i\in\{\textsc{static},\textsc{movable\_static},\textsc{moving}\}$.
PhysVLMM~\cite{ghosh2026physvlmm} adds to each door and drawer its joint (its geometry $\mathcal{G}_i$: joint type $m_i$,
axis $\hat{a}_i$), its group $g$ of look-alike parts, and a first estimate of its forces from the measured look-alikes.
\mba adds the belief $b_{t,i}$ defined below to every element that the map classes as movable and to every door and drawer
($y_i\neq\textsc{static}$), and leaves the other fields unchanged.

\paragraph{Hidden state.}
Object $i$ has the hidden state $s_i=(\psi_i,\mathbf{x}_i,\chi_i)$:
$\psi_i=\log\lambda_i$, where $\lambda_i$ [s$^{-1}$] is the object's own rate of being picked up (the slow rate of
\eqref{eq:renewal});
$\mathbf{x}_i=(x^{o}_i,x^{c}_i)=(\log F^{o}_i,\log F^{c}_i)$, where $F^{o}_i$ is the force needed to move it in the
opening direction and $F^{c}_i$ the largest constant load in the closing direction that it holds still (its support);
and $\chi_i\in\{0,1\}$ indicates a blocked part that cannot reach the target opening.

\paragraph{Actions.}
\mba has five choices. \emph{Look again}: inspect the place where an object was last seen. \emph{Nudge}: push with a
light constant force $f_n$ for a time $T_n$, to test the force. \emph{Open}: pull along the opening direction
$\mathbf{d}_i$, given by the geometry $\mathcal{G}_i$, with a force of at most $F_{\max}$ (controller
\eqref{eq:action}). \emph{Brace}: lean on the open part with a constant load $F_L$ in the closing direction.
\emph{Decline}: do nothing. Nudge, pull and brace act along the part's current direction $\mathbf{d}_i(q)$ at joint
position $q$.

\paragraph{Observations.}
\mba uses three streams.
(a)~\emph{Watching people}: the map's motion record flags element $i$ as moving when people move it; the start
of a flagged motion is a pick-up and its end, at a new position $\mathbf{p}_i$, a placement. These give the completed
rest gaps $r_{i,1},\dots,r_{i,J_i}$ (from a placement to the next pick-up) and the time $a_i$ since the last placement.
(b)~\emph{The robot's touch}: each force action reveals an event. A nudge with force $f_n$ is modelled as revealing
$\one[x^{o}_i\le\log f_n]$, a brace with load $F_L$ reveals $\one[x^{c}_i>\log F_L]$, both threshold events on one
coordinate, and a pull with force limit $F_{\max}$ reveals the joint event $\one[\chi_i=0,\ x^{o}_i\le\log F_{\max}]$.
(c)~\emph{Images} $I_t$ from a camera moved by the robot.

\paragraph{Belief.}
The belief over object $i$ given the history $\mathcal{H}_t$ is
\begin{equation}
\begin{split}
b_{t,i}(\psi,\mathbf{x},\chi)&=p(\psi\mid\mathcal{H}_t)\;p(\mathbf{x},\chi\mid\mathcal{H}_t),\\
p(\mathbf{x},\chi\mid\mathcal{H}_0)&=p(\mathbf{x})\,p(\chi).
\end{split}
\label{eq:belief}
\end{equation}
The belief has two independent parts: how often people move the object ($\psi$), learned by watching them, and its
forces and blocked state ($\mathbf{x},\chi$), learned from the map and by touch. Independence is an assumption: in our
data, objects that people move often are not detectably lighter (Appendix~\ref{sec:s-datacheck}).

\paragraph{Task and utility.}
The task in words, $\mathcal{T}$, sets the utility $U_{\mathcal{T}}$, which the experiments call the \emph{score}: $r_o$
for an opened part, $r_f$ for a failed pull, $r_b$ for a held brace, $r_s$ for a slipped brace, $-c_n$ for a nudge,
$-c_r$ for a look, and $-1$ for acting on an object that is no longer where it was seen (values in \cref{tab:setup}).
These costs set the risk: a brace pays iff it holds with probability above $-r_s/(r_b-r_s)$, which is $0.75$ here.
Language understanding itself is not studied here.

\paragraph{Problem.}
Given the belief $b_t$ and the task $\mathcal{T}$, choose actions that maximise $\E[\sum U_{\mathcal{T}}\mid b_t]$, and,
during every force action, estimate the camera's motion between consecutive images; we score its rotation $\mathbf{R}_t$.

\subsection{Architecture, step by step}
\label{sec:example}
\Cref{fig:architecture}a shows the modules: the map (M1, M2), the still-in-place model (M3), the belief (M4), the
planner (M5), the action (M6) and the tracker used while acting (M7). We follow one real part through them, the middle
of three look-alike drawers in one Hoi! location (orange), for the task ``open it and lean on it'', with
$F_{\max}=13.7$\,N and $F_L=f_n=6.8$\,N.
\textbf{(i)}~\textbf{Map.} The map gives the drawer's label, joint and look-alikes; the other two were measured at $9.9$ and
$8.8$\,N.
\textbf{(ii)}~\textbf{Belief.} From them \mba puts the force to open it near $9.8$\,N (median), against $12.5$\,N for
drawers in general \eqref{eq:pred}. Through the learned link \eqref{eq:mix}, a drawer that turns out harder to open than predicted is also likelier to hold a load.
\textbf{(iii)}~\textbf{Plan.} A pull would likely open it, and a brace is just short of the brace threshold. A nudge could
push the brace past it, but nudging first is worth slightly less than acting now \eqref{eq:vbrace}--\eqref{eq:vnudge}, so \mba does not nudge (values in \cref{fig:architecture}b).
\textbf{(iv)}~\textbf{What a nudge would tell.} Had the drawer not moved, it would need more than $f_n$: a pull would be less likely to open it
and, through the link, a brace would pass the threshold \eqref{eq:touch}.
\textbf{(v)}~\textbf{Act.} Opening pays \eqref{eq:qopen}, so \mba pulls while tracking the camera
(\cref{sec:camera}); the drawer opens at $8.6$\,N, so the score is $1.0$. \mba does not brace, although the drawer holds up to $11.5$\,N; without the link it would brace (score $2.0$). In the main setting, this drawer and one look-alike are the only parts on which the link changes a decision (\cref{sec:r-pulls}).
\textbf{(vi)}~\textbf{Still in place?} For an object that people move, such as a colander in a kitchen (M3), \mba first
asks whether it is still there \eqref{eq:presence}. Just after the colander is put down, \mba gives $0.68$ that it is
still in place two minutes later; after a minute at rest, the same chance is $0.93$, as a quick pick-up is now unlikely. \mba looks again
when this chance is too low \eqref{eq:relook}.
This part shows every step, although \mba does not execute the nudge here; over all parts the nudge and the link change little (\cref{sec:r-pulls}).
\Cref{alg:mba} lists the steps; \cref{sec:algorithm} gives their equations.

\subsection{Belief and decisions}
\label{sec:algorithm}

\subsubsection{Belief from the map}
\label{sec:belief}
This gives steps (ii), (iv) and (vi). All parameters below are fitted on training data only, by (marginal) maximum
likelihood with weak priors or, for $\varrho_k$, by resampling locations (Appendix~\ref{sec:s-params}).

\paragraph{Force prior.}
For a direction $d\in\{o,c\}$, the map prior is PhysVLMM's hierarchical model~\cite{ghosh2026physvlmm} on
look-alike groups,
\begin{equation}
x^{d}_i\mid\mu^{d}_{g}\sim\N(\mu^{d}_{g},\tau_d^2),\qquad \mu^{d}_{g}\sim\N(\mu^{d}_{m},\sigma_d^2),
\label{eq:p4}
\end{equation}
where $\mu^{d}_m$ is the training mean of $x^{d}$ for joint type $m$. Given the forces $x^{d}_j$ measured on the set
$\mathcal{L}_i$ of $n$ look-alikes of part $i$ (part $i$ excluded), the predictive is $\N(\hat{x}^{d}_i,v^{d}_i)$ with
\begin{equation}
\hat{x}^{d}_i=\frac{\mu^{d}_m/\sigma_d^2+\sum_{j\in\mathcal{L}_i}x^{d}_j/\tau_d^2}{1/\sigma_d^2+n/\tau_d^2},\qquad
v^{d}_i=\tau_d^2+\Big(\frac{1}{\sigma_d^2}+\frac{n}{\tau_d^2}\Big)^{-1}.
\label{eq:pred}
\end{equation}
A touch by the robot updates only the touched part \eqref{eq:touch}.

\paragraph{Learned link between the forces.}
Let $\tilde{x}^{d}=(x^{d}-\hat{x}^{d})/\sqrt{v^{d}}$ be the standardised errors of \eqref{eq:pred}. The pair
$(\tilde{x}^{o},\tilde{x}^{c})$ is bivariate normal with correlation $\varrho$, the \emph{link}; its estimate
$\hat\varrho$ is the correlation of all training errors. To keep its uncertainty, \mba resamples training locations,
which gives samples $\varrho_1,\dots,\varrho_K$, and uses the mixture
\begin{equation}
p(\mathbf{x}_i)=\frac{1}{K}\sum_{k=1}^{K}\N\!\big(\mathbf{x}_i;\,\hat{\mathbf{x}}_i,\,\Sigma_i(\varrho_k)\big),
\label{eq:mix}
\end{equation}
where $\Sigma_i(\varrho)$ has diagonal $(v^{o}_i,v^{c}_i)$ and off-diagonal $\varrho\sqrt{v^{o}_iv^{c}_i}$.
With $\varrho_k\approx0$ the forces are independent. All components have the same marginals, so observing $x^{o}$ alone
leaves their weights unchanged (Appendix~\ref{sec:s-params}).

\paragraph{Touch update.}
A nudge or brace outcome is a threshold event $z$ on one force, so
\begin{equation}
p(\mathbf{x}_i\mid\mathcal{H},z)\propto p(\mathbf{x}_i\mid\mathcal{H})\,\one[z],
\label{eq:touch}
\end{equation}
computed on a $G\times G$ grid of $(x^{o},x^{c})$, each cell weighted by its share on the event's side
(Appendix~\ref{sec:s-params}). These outcomes depend on $\mathbf{x}$ only, so the blocked state stays independent of the forces.
A failed pull couples them: it reveals $\{\chi=1\}\cup\{x^{o}>\log F_{\max}\}$, so
$\Pb(\chi=1\mid\text{fail})=\beta/(1-\bar p_o)$, where $\beta=\Pb(\chi=1)$ and
$\bar p_o=(1-\beta)\,\Pb(x^{o}\le\log F_{\max})$ is the probability that a pull opens the part. The blocked prior is
$\beta=(N_b+1)/(N+2)$ for $N_b$ blocked of the $N$ training parts of the same kind (\eg door or drawer); the real pulls have
no blocked label, so $\beta=0$ there.

\paragraph{Still-in-place model.}
The rest gap $R$ between a placement and the next pick-up has the survival function
\begin{equation}
S(r\mid \psi)=w_f\,e^{-\lambda_f r}+(1-w_f)\,e^{-e^{\psi}r}.
\label{eq:renewal}
\end{equation}
The quick pick-up, with weight $w_f$ and rate $\lambda_f$, is shared by all objects. The slow rate $e^{\psi}=\lambda_i$
is the object's own, with $\psi_i\sim\N(\mu_{\ell},\omega^2)$ for objects with label $\ell$ and
$\mu_{\ell}\sim\N(\nu_{h},\eta^2)$ for labels of category $h$. With the density $f=-\partial_r S$ and the prior
$p(\psi)$ of $\psi_i$, Bayes' rule over $\psi$ gives the probability that object $i$ is still in place after a further
time $\Delta$ (derivation and fitting in Appendix~\ref{sec:s-params}),
\begin{equation}
P_i(\Delta)=\frac{\int S(a_i+\Delta\mid \psi)\prod_{j=1}^{J_i}f(r_{i,j}\mid \psi)\,p(\psi)\,d\psi}
{\int S(a_i\mid \psi)\prod_{j=1}^{J_i}f(r_{i,j}\mid \psi)\,p(\psi)\,d\psi}.
\label{eq:presence}
\end{equation}

\subsubsection{Decisions}
\label{sec:decision}

\paragraph{Planner.}
This gives steps (iii) and (v). For a part with force belief $W$ (the grid of \eqref{eq:mix}--\eqref{eq:touch}), the
values of bracing, opening and nudging first are
\begin{align}
V_B(W)&=\max\{0,\;p_b r_b+(1-p_b)r_s\},\label{eq:vbrace}\\
V_O(W)&=\max\{0,\;Q_O(W)\},\label{eq:vopen}\\
Q_O(W)&=\bar p_o\big[r_o+V_B(W_{\le})\big]+(1-\bar p_o)r_f,\label{eq:qopen}\\
V_N(W)&=-c_n+p_n V_O(W_{\mathrm{mv}})+(1-p_n)V_O(W_{\mathrm{st}}),\label{eq:vnudge}
\end{align}
with $p_b=\Pb_W(x^{c}>\log F_L)$ the probability that a brace holds, $p_n=\Pb_W(x^{o}\le\log f_n)$ that a nudge moves
the part, $\bar p_o$ as above under $W$, and $W_{\le}$, $W_{\mathrm{mv}}$, $W_{\mathrm{st}}$ the beliefs after the
events \emph{opened}, \emph{moved} and \emph{stayed} \eqref{eq:touch}. \mba nudges iff $V_N>V_O$, opens iff $Q_O>0$,
braces iff $p_b r_b+(1-p_b)r_s>0$, and otherwise declines. We assume that a blocked part still moves under a nudge, so a
nudge leaves $\beta$ unchanged. Each of the three stages has two outcomes, so backward induction is exact on the grid,
in $O(G^2)$ time. When bracing adds nothing and $\beta=0$, the open rule is PhysVLMM's risk
query~\cite{ghosh2026physvlmm} with its risk level set to $\delta=r_o/(r_o-r_f)$.
For an object that may have moved, $Q_O$ becomes $P_iQ_O-(1-P_i)$, with $P_i=P_i(\Delta)$ from \eqref{eq:presence} at
the time $\Delta$ of the action. For an object that will serve as a landmark or a place to put things back, \mba looks
again iff the expected loss of acting on it unseen exceeds the cost of a look,
\begin{equation}
1-P_i(\Delta)>c_r .
\label{eq:relook}
\end{equation}

\paragraph{One nudge, opposite decisions.}
\begin{proposition}
\label{prop:opposite}
Let the nudge outcome $z$ depend on $(\mathbf{x},\chi)$ only through $x^{o}$, with $\Pb(z=\text{moved}\mid x^{o})$
non-increasing in $x^{o}$ (a threshold event is a special case). Then, for any prior on $x^{o}$,
\[
\bar p_o(\cdot\mid\text{moved})\;\ge\;\bar p_o\;\ge\;\bar p_o(\cdot\mid\text{stayed}).
\]
If every component of \eqref{eq:mix} has $\varrho_k\ge0$, the inequalities for $p_b$, before or after the opening
update, point the other way.
\end{proposition}
This is a monotone likelihood ratio argument~\cite{milgrom1981good} (all proofs in Appendix~\ref{sec:s-method}); a few fitted $\varrho_k$ are slightly negative, and
the check on the actual belief is in Appendix~\ref{sec:s-props}. So a nudge that moves the part makes opening more attractive
and bracing less safe at once; a belief that does not link the two forces would miss the second effect.

\paragraph{Value of the link.}
\begin{proposition}
\label{prop:bound}
Consider a decision that gains $r_+>0$ if event $A$ holds and $r_-<0$ otherwise, taken iff $\Pb(A)>p^{\star}=-r_-/(r_+-r_-)$.
Let an observation $Z$ give $p=\Pb(A\mid Z)$ with mean $p_0$, variance $\sigma_p^2$ and $\kappa=|p_0-p^{\star}|$. Then the
value of information of $Z$, the gain in expected utility from deciding after observing it, obeys
\begin{equation}
\mathrm{VOI}(Z)\le(r_+-r_-)\,\frac{\sqrt{\sigma_p^2+\kappa^2}-\kappa}{2}.
\label{eq:bound}
\end{equation}
If $A=\{\tilde{x}^{c}>\theta\}$, with $\theta$ the standardised $\log F_L$, and $Z$ reveals $\tilde{x}^{o}$ with
$\operatorname{corr}(\tilde{x}^{o},\tilde{x}^{c})=\varrho$, then $\sigma_p^2=\Phi_2(-\theta,-\theta;\varrho^2)-\Phi(-\theta)^2\le\arcsin(\varrho^2)/(2\pi)$
($\Phi$, $\Phi_2$: standard normal and bivariate normal CDFs), and any function of $\tilde{x}^{o}$
(\eg a nudge) has no larger $\sigma_p^2$.
\end{proposition}
The bound is Scarf's and Lo's option-price inequality~\cite{scarf1958minmax,lo1987semiparametric} applied to the value
of information. It is cheap to evaluate, so it can be checked for each part before planning. For the brace decision,
evaluated before any touch, it bounds what ignoring the link can lose; evaluated after the part opened, it bounds what a
further reading of the opening force, such as a nudge, can add.

\subsubsection{Tracking while acting}
\label{sec:camera}

This runs during step (v). The camera is rigidly mounted on the gripper and moves by $(\mathbf{R}_t,\mathbf{t}_t)$ between images. A static world
point with normalised coordinates $\bar{\mathbf{u}}\mapsto\bar{\mathbf{u}}'$ satisfies
$\bar{\mathbf{u}}'^{\top}E\,\bar{\mathbf{u}}=0$, $E=[\mathbf{t}_t]_{\times}\mathbf{R}_t$. While the robot holds part $i$,
the part moves with the gripper, so its points (and the closed fingers) keep their camera coordinates and their pixels do
not move,
\begin{equation}
\mathbf{u}'_j=\mathbf{u}_j+\boldsymbol{\epsilon}_j,\qquad \boldsymbol{\epsilon}_j\sim\N(\mathbf{0},\sigma_{co}^2\mathbf{I}),
\label{eq:rigid}
\end{equation}
up to tracking noise and grasp slip.

\begin{proposition}
\label{prop:degenerate}
A feature obeying \eqref{eq:rigid} with $\boldsymbol{\epsilon}_j=\mathbf{0}$ satisfies the epipolar constraint of every
motion with $\mathbf{R}=\mathbf{I}$, whatever $\mathbf{t}$. If such features outnumber the consensus set of the true motion,
a hypothesis with $\mathbf{R}=\mathbf{I}$ has the larger consensus, and its rotation error equals the true rotation angle.
\end{proposition}

\cref{prop:degenerate} is the classical zero-flow degeneracy (numerical check in Appendix~\ref{sec:s-degeneracy}), so the held
part's features must be removed \emph{before} $E$ is estimated. \mba finds them by their near-zero flow
\eqref{eq:rigid} and by an \emph{action prior}: where, for the current grasp state and part type, the held part appeared
in training sessions. It then estimates $E$ by RANSAC on the remaining features (Appendix~\ref{sec:s-method}) and scores only
its rotation.

\subsection{Algorithm}
\label{sec:alg}
\Cref{alg:mba} gathers the steps of \cref{sec:example} for one part.
\begin{algorithm}[t]
\caption{\mba for one part and one task}
\label{alg:mba}
\begin{algorithmic}[1]
\Require map, part $i$, task $\mathcal{T}$ (utility $U_{\mathcal{T}}$), settings $f_n,F_{\max},F_L$
\State $(\hat{x}^{d}_i,v^{d}_i)\gets$ predictive from look-alikes \eqref{eq:pred}, $d\in\{o,c\}$
\State $W\gets$ grid of the mixture \eqref{eq:mix} with the learned $\varrho_{1..K}$; $\beta\gets$ blocked prior of part $i$'s kind
\If{$V_N(W)>V_O(W)$} \Comment{\eqref{eq:vbrace}--\eqref{eq:vnudge}}
  \State nudge with constant force $f_n$ along $\mathbf{d}_i$; observe moved/stayed
  \State $W\gets$ update \eqref{eq:touch}
\EndIf
\If{$Q_O(W)\le0$ \eqref{eq:qopen}} \Return decline \EndIf
\State pull along $\mathbf{d}_i$ with limit $F_{\max}$ \eqref{eq:action}; track camera
\If{not opened} \Return failed (update map) \EndIf
\State $W\gets$ update with $x^{o}\le\log F_{\max}$ \eqref{eq:touch}
\If{$V_B(W)>0$} brace with $F_L$ along $-\mathbf{d}_i$; observe held/slipped \EndIf
\State write the outcomes to part $i$'s map entry \eqref{eq:touch}
\end{algorithmic}
\end{algorithm}

\section{Results and discussions}
\label{sec:results}

\mba is tested on four questions; \Cref{tab:setup} gives their data, ground truth and scoring. Besides
published methods, \mba is compared with ablations (abl.), which remove one of its parts, and with controls (ctrl.). All methods get the same inputs: the datasets' annotations stand in for the map, except that the real pulls use
PhysVLMM's part types and look-alike groups predicted from the scans. Each location or kitchen is tested
once, with every parameter fitted on the others; in simulation the test layouts are not used for training. Published methods run from their authors' code, except PhysVLMM's risk query (no released code), which we
re-implement from its paper on \mba's map prior. Differences are paired, with $95\,\%$ intervals from a bootstrap over locations or
kitchens; bold: the interval excludes $0$. Settings and measures were fixed before scoring, except the simulation's main setting, chosen after a pilot on two test kitchens (Appendix~\ref{sec:s-details}), its damage measure, fixed
during the SmolVLA runs and before any Octo result, its controls, which drop the brace, and its class-average-prior
planner, both chosen after runs with a forced brace, the hyper-prior on category means in the still-in-place model, added
after one fold diverged, and the comparisons with the published risk query and with Perpetua and the exclusion of $20$ mislabelled scan instances (Appendix~\ref{sec:s-details}), added
after the main results.

\begin{table*}[!t]
\centering\small
\setlength{\tabcolsep}{4pt}
\caption{\textbf{The four questions, their ground truth and scoring.} Score: $+1$ opened, $-1$ failed pull, $+1$ held
brace, $-3$ slipped brace, $-0.1$ per nudge; \emph{unsafe}: failed pulls and slipped braces. Real pulls:
Law-to-Action~\cite{mabrok2026l2a}, TANav~\cite{li2026tanav} and MessyMem~\cite{banwasi2026messymem} have released no
code, and Phys2Real~\cite{wang2026phys2real} needs its own simulator and robot arm.}
\label{tab:setup}
\resizebox{\textwidth}{!}{%
\begin{tabular}{>{\raggedright\arraybackslash}p{3.4cm}>{\raggedright\arraybackslash}p{3.9cm}>{\raggedright\arraybackslash}p{6.4cm}>{\raggedright\arraybackslash}p{4.0cm}>{\raggedright\arraybackslash}p{3.0cm}}
\toprule
Question & Data and test set & Ground truth & Measures and settings & Published methods \\
\midrule
\textbf{Real pulls}: do the map's force estimates improve pulling decisions? & Hoi!~\cite{engelbracht2026hoi}: $271$
pulls by a person with a force-sensing gripper ($100$\,Hz) on $143$ parts; the $121$ parts with both directions and a map entry, in $17$ locations &
read from the recorded human pull; no robot acts.\newline
\emph{Open}: peak force in the first half of the opening.\newline
\emph{Hold}: peak of a closing pull (Hoi! has no holding trials).\newline
\emph{Nudge}: did the part move before the force reached the nudge force (differs from the planner's model on $43$
parts in the main setting).\newline
\emph{Blocked}: none ($\beta=0$). &
score, unsafe; force limit: training locations' median force (main), fixed $8$\,N, sweep of $420$ settings &
PhysVLMM's risk query~\cite{ghosh2026physvlmm} (re-implemented) \\
\textbf{Kitchen activity}: is the object still in place? & HD-EPIC~\cite{perrett2025hdepic}: $156$ videos, $8{,}382$
objects, $19{,}620$ moves; $309{,}588$ queries in $9$ kitchens &
annotated pick-ups: still in place if none starts within $\Delta$ &
Brier score, log-loss, re-look cost (extra cost over knowing the answer); $\Delta=30$\,s, $2$\,min, $10$\,min; look
cost $0.05$--$0.5$ &
persistence filter~\cite{rosen2016persistence}, FreMEn~\cite{krajnik2017fremen}, Perpetua~\cite{saavedra2025perpetua} \\
\textbf{Simulation}: does \mba help vision-language-action policies at a safe force limit? &
RoboCasa~\cite{nasiriany2024robocasa} in ManiSkill3~\cite{tao2025maniskill3}, physics fitted to
Hoi!~\cite{ghosh2026physvlmm}: $120$ kitchens; $1{,}433$ parts in $36$ kitchens &
the executed episode: optional nudge, opening to half range within $6$\,s, optional brace for $1$\,s, by a
force-limited controller at the handle (no arm or grasp) &
score, opened, declined, damage (share of parts pushed above the safe limit for $\ge0.5$\,s); safe limit: training parts' median force (main), robot limit
$30$\,N &
SmolVLA~\cite{shukor2025smolvla}, Octo~\cite{octo2024}, PhysVLMM's risk query (re-implemented) \\
\textbf{Wrist camera}: does the action prior help tracking? & Hoi! gripper camera ($10$ frames/s), one session in each of
$18$ locations (the $17$ above and one more); $2{,}702$ frame pairs while holding a part &
the recording device's own camera trajectory &
failed pair: rotation error $>2^\circ$ or no pose; each SLAM system run three times &
ORB-SLAM2~\cite{murartal2017orbslam2}, DynaSLAM~\cite{bescos2018dynaslam} \\
\bottomrule
\end{tabular}}
\end{table*}

\begin{figure*}[!t]
\centering
\includegraphics[width=\textwidth]{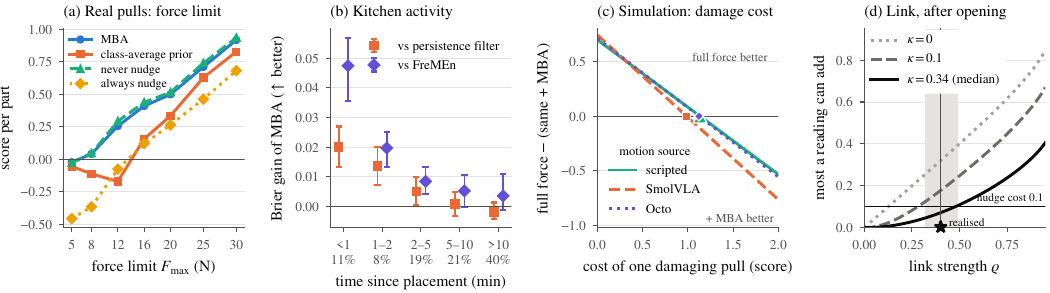}
\vspace{-1.5em}
\caption{\textbf{Results.} (a) Real pulls, mean over the sweep settings (real nudge outcome). (b) Kitchen activity:
$95\,\%$ intervals; percentages: share of queries per bin. (c) Simulation at $30$\,N; markers: break-even (SmolVLA:
episodes with a force trace). (d) Bound of \cref{prop:bound} on what a reading of the opening force adds to the brace
decision after the part opened ($r_+-r_-=4$), for parts at three distances $\kappa$ of the prior from the brace threshold; band: learned $\hat\varrho$ over the folds (line: median); horizontal line: nudge cost; star: realised value.}
\label{fig:results}
\end{figure*}

\subsection{Real pulls}
\label{sec:r-pulls}

\begin{figure*}[t]
\centering
\includegraphics[width=\textwidth]{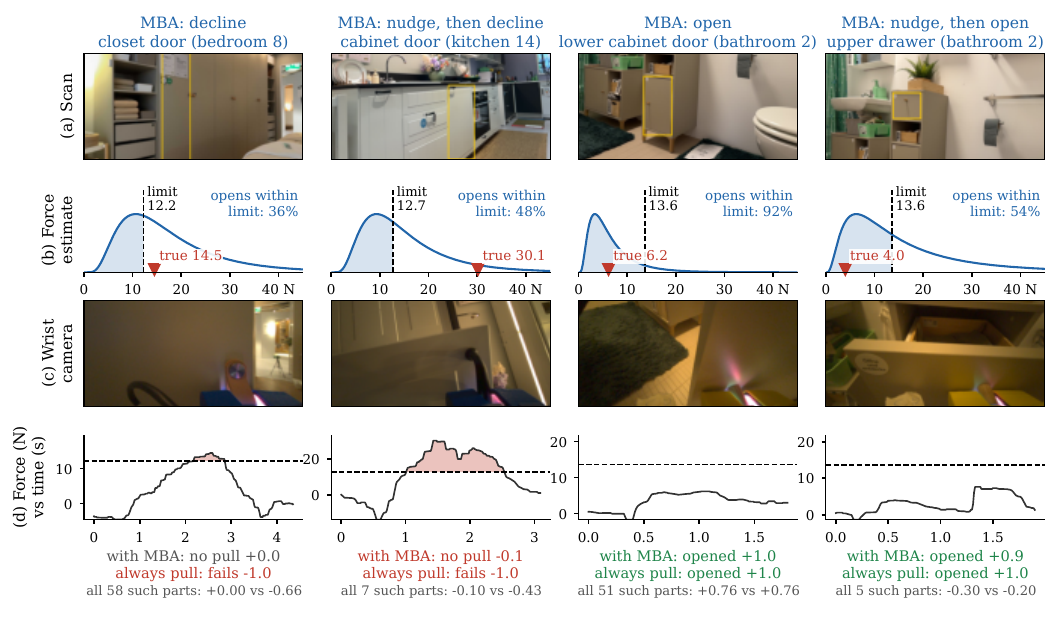}
\vspace{-1.5em}
\caption{\textbf{Real pulls, start to finish.} One test part for each of the four action sequences \mba chooses on real pulls (in the main setting it braces no part). Dashed: the
location's force limit; red $\blacktriangledown$: true force; shaded: force above the limit, where a pull at the limit
fails. Below: score with \mba and with always pulling at the limit (our control), and their means over all test parts
with the same decision.}
\label{fig:real}
\end{figure*}

\begin{table*}[!t]
\centering\small
\setlength{\tabcolsep}{5pt}
\caption{\textbf{Real pulls.} Simulated nudge: the nudge outcome read as the opening force, as the planner models it.
Sweep: settings in which \mba is better / worse (about $10$ per side are expected by chance; Appendix~\ref{sec:s-details} and \cref{fig:s-sweep}). Rows below the line are
ablations.}
\label{tab:touch}
\resizebox{\textwidth}{!}{%
\begin{tabular}{lcccccc}
\toprule
& \multicolumn{3}{c}{Training locations' median force} & \multicolumn{2}{c}{\mba\ $-$ method, other conditions} & \\
\cmidrule(lr){2-4}\cmidrule(lr){5-6}
Method & Score & Unsafe & \mba\ $-$ method & Simulated nudge & Fixed $8$\,N & Sweep \\
\midrule
\mba & 0.304 & 7.4\,\% & -- & -- & -- & -- \\
PhysVLMM risk query~\cite{ghosh2026physvlmm}, $\delta=0.1$ & 0.083 & 3.3\,\% & $\mathbf{+0.221}$ {\scriptsize$[+0.043,+0.450]$} & $\mathbf{+0.246}$ {\scriptsize$[+0.073,+0.469]$} & $+0.057$ {\scriptsize$[-0.001,+0.128]$} & 258/10 \\
PhysVLMM risk query, $\delta=0.5$ & 0.322 & 7.4\,\% & $-0.018$ {\scriptsize$[-0.059,+0.007]$} & $+0.007$ {\scriptsize$[-0.023,+0.037]$} & $+0.007$ {\scriptsize$[0.000,+0.021]$} & 24/42 \\
\midrule
Class-average prior & $-0.075$ & 25.6\,\% & $\mathbf{+0.379}$ {\scriptsize$[+0.240,+0.529]$} & $\mathbf{+0.247}$ {\scriptsize$[+0.120,+0.419]$} & $+0.049$ {\scriptsize$[-0.007,+0.117]$} & 209/5 \\
Never nudge & 0.322 & 7.4\,\% & $-0.018$ {\scriptsize$[-0.059,+0.007]$} & $+0.007$ {\scriptsize$[-0.023,+0.037]$} & $+0.007$ {\scriptsize$[0.000,+0.021]$} & 0/62 \\
Always nudge & 0.065 & 28.9\,\% & $\mathbf{+0.239}$ {\scriptsize$[+0.103,+0.388]$} & $+0.049$ {\scriptsize$[-0.011,+0.086]$} & $\mathbf{+0.355}$ {\scriptsize$[+0.207,+0.499]$} & 394/0 \\
No link ($\varrho=0$) & 0.321 & 7.4\,\% & $-0.017$ {\scriptsize$[-0.059,0.000]$} & $-0.017$ {\scriptsize$[-0.059,0.000]$} & $+0.032$ {\scriptsize$[-0.002,+0.100]$} & 1/1 \\
Perfect link ($\varrho=1$) & 0.261 & 9.9\,\% & $+0.043$ {\scriptsize$[-0.081,+0.225]$} & $+0.035$ {\scriptsize$[-0.089,+0.219]$} & $-0.005$ {\scriptsize$[-0.017,0.000]$} & 36/0 \\
\bottomrule
\end{tabular}}
\end{table*}

The map's force estimates matter most (\cref{tab:touch}): with the class-average prior instead, the score drops at every force
limit (\cref{fig:results}a) and in all locations but one (\cref{fig:cluster}). PhysVLMM's own risk query pulls only when
it is at least $1-\delta$ sure that the part opens, and has no query for bracing. At its published $\delta=0.1$ it is
more cautious than \mba: it fails fewer pulls but opens fewer parts, and \mba scores higher, significantly in the main setting (not before the exclusion, Appendix~\ref{sec:s-details}) and in more than half of the sweep settings. With $\delta=0.5$, the risk at which a pull pays under these
costs, it ties \mba in the main setting, as expected. Under these costs \mba's open rule is this query whenever bracing adds nothing (every
part here), so the two differ only through the nudge and the brace; over the sweep it is significantly ahead in $42$ settings and behind in $24$. Against the published $\delta$, \mba's gain is
therefore that it sets its risk from the task's costs, not a better estimate (also under other costs, Appendix~\ref{sec:supp-results}). With the real nudge outcome, not nudging scores slightly higher in the main setting (not significantly) and significantly higher in $62$ sweep settings, never significantly lower. A real nudge senses the force at which motion starts: for opening, a median $70\,\%$ of the peak
that decides the pull, and below half of it on $45\,\%$ of parts. The link between opening and closing force is real ($\hat\varrho=0.32$--$0.49$ over the folds) but
in the main setting changes only $2$ decisions, both braces that would have held, and no score significantly. It changes few because the link reaches the brace decision mainly through the opening: opening flips it on $23$ parts, but \mba declines $21$ of them (chance of opening at most $0.30$). After the part has opened, \cref{prop:bound} bounds what a further reading of the opening force, such as a
nudge, can add to the brace decision (\cref{fig:results}d): at $\hat\varrho\approx0.40$ this bound exceeds the nudge cost only for parts
whose probability that a brace holds is within $0.23$ below and $0.08$ above the brace threshold, and only $33$ parts are. \Cref{fig:real} follows, for each decision, the first part
in location order with a wrist-camera recording and a scan box (one skipped: a QR-code sheet covers its scan): \mba gains on the parts it declines that would fail, and on the parts it opens
it scores as pulling would, less any nudge cost.

\begin{figure}[t]
\centering
\includegraphics[width=\columnwidth]{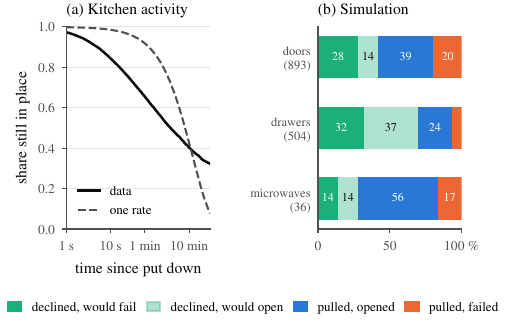}
\vspace{-1.5em}
\caption{\textbf{What \mba decides.} (a) Kitchen activity: share of objects still in place after being put down (all
rest gaps), and a single rate fitted to the same gaps. (b) Simulation, safe limit: \mba's decision on each test part (in
\%), and whether the part opens within the limit.}
\label{fig:examples}
\end{figure}

\subsection{Kitchen activity}
\label{sec:r-presence}

\mba has the lowest Brier score, log-loss and re-look cost (\cref{tab:presence}). Its re-look cost is significantly
lower than all three methods'. On Brier score it is ahead of FreMEn and Perpetua in every kitchen and of the persistence filter in all but two
(\cref{fig:cluster}), but its lead over the persistence filter is narrow (paired $t$-test $p=0.039$; the $95\,\%$
interval includes $0$). Perpetua's output follows only its slowest lifetime, so it misses the quick pick-up. The gain comes from the quick pick-up: about a quarter of placements end in a pick-up within
$30$\,s, far more than a single rate predicts (\cref{fig:examples}a), so the gain is largest just after a placement
(\cref{fig:results}b); without it \mba loses its lead. It is less well calibrated than the persistence filter and FreMEn (\cref{fig:s-exp3}).

\begin{table}[!tb]
\centering\small
\setlength{\tabcolsep}{3pt}
\caption{\textbf{Kitchen activity.} Re-look cost: mean over the look costs and the three horizons. Calib.: calibration
error over ten bins. Without object history: the object's own past gaps are ignored.}
\label{tab:presence}
\resizebox{\columnwidth}{!}{%
\begin{tabular}{lccccl}
\toprule
Method & Brier\,$\downarrow$ & Log-loss\,$\downarrow$ & Re-look cost\,$\downarrow$ & Calib.\,$\downarrow$ & Re-look cost $-$ \mba \\
\midrule
\mba & .0762 & .257 & .0472 & .026 & -- \\
Persistence filter~\cite{rosen2016persistence} & .0811 & .273 & .0522 & .011 & $\mathbf{+.0049}$ {\scriptsize$[+.0005,+.0104]$} \\
FreMEn~\cite{krajnik2017fremen} & .0905 & .310 & .0603 & .021 & $\mathbf{+.0131}$ {\scriptsize$[+.0082,+.0186]$} \\
Perpetua~\cite{saavedra2025perpetua} & .0963 & .366 & .0705 & .069 & $\mathbf{+.0233}$ {\scriptsize$[+.0140,+.0388]$} \\
\midrule
\mba without quick pick-up & .1005 & .334 & .0576 & .097 & $\mathbf{+.0104}$ {\scriptsize$[+.0067,+.0125]$} \\
\mba without object history & .0777 & .260 & .0488 & .022 & $+.0016$ {\scriptsize$[-.0008,+.0054]$} \\
Time since placement (ctrl.) & .0811 & .274 & .0519 & .003 & $\mathbf{+.0047}$ {\scriptsize$[+.0003,+.0100]$} \\
Label rate (ctrl.) & .1017 & .329 & .0652 & .096 & $\mathbf{+.0180}$ {\scriptsize$[+.0108,+.0236]$} \\
Global rate (ctrl.) & .1406 & .422 & .0887 & .156 & $\mathbf{+.0415}$ {\scriptsize$[+.0335,+.0458]$} \\
\bottomrule
\end{tabular}}
\end{table}

\subsection{Simulation}
\label{sec:r-closed}

\begin{table}[!tb]
\centering\small
\setlength{\tabcolsep}{3pt}
\caption{\textbf{Simulation} (main setting). \mba\ $-$ method: against the same motion source. The score charges nothing
for damage (\cref{fig:results}c adds its cost). $^{a}$On the $1{,}076$ of $1{,}433$ SmolVLA episodes with a recorded
force trace (recorded only after the damage measure was fixed). $^{b}$Scored on the safe-limit episodes; a part it does not pull scores $0$.}
\label{tab:closed}
\resizebox{\columnwidth}{!}{%
\begin{tabular}{lccccl}
\toprule
Method & Score & Opened & Declined & Damage & \mba\ $-$ method \\
\midrule
SmolVLA + \mba & 0.165 & 33\,\% & 51\,\% & 0$^{a}$ & -- \\
SmolVLA~\cite{shukor2025smolvla}, safe limit & 0.101 & 55\,\% & 0 & 0$^{a}$ & $\mathbf{+0.064}$ {\scriptsize$[+0.021,+0.107]$} \\
SmolVLA, full $30$\,N & 0.866 & 93\,\% & 0 & 75\,\%$^{a}$ & see \cref{fig:results}c \\
\midrule
Octo + \mba & 0.188 & 34\,\% & 51\,\% & 0 & -- \\
Octo~\cite{octo2024}, safe limit & 0.144 & 57\,\% & 0 & 0 & $\mathbf{+0.043}$ {\scriptsize$[+0.001,+0.088]$} \\
Octo, full $30$\,N & 0.898 & 95\,\% & 0 & 63\,\% & see \cref{fig:results}c \\
\midrule
\mba (scripted motion) & 0.176 & 34\,\% & 51\,\% & 0 & -- \\
\mba without nudge & 0.166 & 33\,\% & 50\,\% & 0 & $+0.010$ {\scriptsize$[-0.002,+0.023]$} \\
Class-average prior only (abl.) & 0.144 & 23\,\% & 75\,\% & 0 & $+0.032$ {\scriptsize$[-0.011,+0.078]$} \\
Scripted, safe limit (ctrl.) & 0.118 & 56\,\% & 0 & 0 & $\mathbf{+0.059}$ {\scriptsize$[+0.016,+0.106]$} \\
PhysVLMM risk query$^{b}$, $\delta=0.1$ & 0.000 & 0.1\,\% & 99.9\,\% & 0 & $\mathbf{+0.176}$ {\scriptsize$[+0.113,+0.239]$} \\
PhysVLMM risk query$^{b}$, $\delta=0.5$ & 0.161 & 37\,\% & 43\,\% & 0 & $+0.015$ {\scriptsize$[-0.005,+0.034]$} \\
\bottomrule
\end{tabular}}
\end{table}

The policies imitate a scripted controller and see no force. At the safe limit (\cref{tab:setup}),
\mba improves both policies (\cref{tab:closed}), mostly by declining pulls it believes would fail; without the two pilot kitchens the gains are $+0.062$ and $+0.041$, the second not significant (Appendix~\ref{sec:supp-results}). With scripted motion, a
planner with only the class-average prior gets $45\,\%$ of this gain, and \mba's lead over it is not significant. PhysVLMM's risk query almost never pulls at $\delta=0.1$, nearly every answer being an abstention (``insufficient evidence''); as on real
pulls, \mba beats it there and ties it at
$\delta=0.5$, also with both policies (Appendix~\ref{sec:supp-results}). At the full
$30$\,N the policies open most parts but damage most of them, so \mba wins once a damaging pull costs more than about $1$
(\cref{fig:results}c). Its declines are mostly right on doors, but on drawers more of the declined parts would have
opened than failed (\cref{fig:examples}b, \cref{tab:kind}). At the fixed $8$\,N limit adding \mba changes nothing
(Appendix~\ref{sec:supp-results}).

\subsection{Wrist camera}
\label{sec:r-camera}

Two-view odometry fails more than twice as often on pairs taken while holding a part as on free pairs ($15.7\,\%$),
and monocular SLAM tracks under a fifth of them. \mba's action prior does not help measurably, because the held part's near-zero image motion
\eqref{eq:rigid} already identifies it. \mba's tracker fails slightly less often than plain two-view odometry, but not significantly
($35.5$ vs $38.3\,\%$; \cref{tab:camera,fig:s-exp2}; run times in Appendix~\ref{sec:s-details}).

\section{Conclusion}
\label{sec:conclusion}

We presented Motion Belief Action, a decision layer that reads a scene map's record of what people move and what the
robot has touched, and chooses each action by the task's costs. On real pulls, its gain over a part-type average comes from the map's force estimates. Its gain over PhysVLMM's risk query at the published risk level comes from setting the risk from the task's costs. In
simulation, the gain comes mostly from declining pulls, and a part-type average gets about half of it. A published risk rule given the same costs does as well in the main settings. The additions beyond this (the link between the forces, the probing nudge and
the tracker's action prior) add nothing measurable there, and over the real-pull sweep the nudge costs a little. For still-in-place prediction, what matters is that people often pick an object up
again soon after putting it down.

\paragraph{Limitations.} The real pulls were made by a person, not a robot, so the nudge and the support are read from
recorded pulls. The costs are set by hand; when a failed pull costs nine times an opening, no tested rule beats not pulling (Appendix~\ref{sec:supp-results}). The simulated robot opens in the true direction, and its link between the
forces comes from the simulator's shared friction, fitted per part kind (Appendix~\ref{sec:s-details}). No object in our data has both people moving it and measured forces,
so a decision combining them is not tested.

{
    \small
    \bibliographystyle{ieeenat_fullname}
    \bibliography{main}

\begin{thebibliography}{43}
\providecommand{\natexlab}[1]{#1}
\providecommand{\url}[1]{\texttt{#1}}
\expandafter\ifx\csname urlstyle\endcsname\relax
  \providecommand{\doi}[1]{doi: #1}\else
  \providecommand{\doi}{doi: \begingroup \urlstyle{rm}\Url}\fi

\bibitem[Adkins et~al.(2026)Adkins, Ravisankar, and Biswas]{adkins2026presist}
Amanda Adkins, Tarunvidyut Ravisankar, and Joydeep Biswas.
\newblock {PreSIST}: Vision-language-informed object persistence prediction in
  open-world scenes.
\newblock \emph{arXiv:2607.04057}, 2026.

\bibitem[Banwasi et~al.(2026)Banwasi, Muckelroy~III, Sundaresan, Zhao, Bohg,
  and Ho]{banwasi2026messymem}
Anuva Banwasi, William Muckelroy~III, Priya Sundaresan, Linfeng Zhao, Jeannette
  Bohg, and Cherie Ho.
\newblock {MessyMem}: Learning-from-doing memory for mobile manipulation.
\newblock In \emph{CoRL}, 2026.
\newblock arXiv:2609.15976.

\bibitem[Bescos et~al.(2018)Bescos, F{\'a}cil, Civera, and
  Neira]{bescos2018dynaslam}
Berta Bescos, Jos{\'e}~M. F{\'a}cil, Javier Civera, and Jos{\'e} Neira.
\newblock {DynaSLAM}: Tracking, mapping, and inpainting in dynamic scenes.
\newblock \emph{IEEE Robotics and Automation Letters}, 3\penalty0 (4):\penalty0
  4076--4083, 2018.

\bibitem[Black et~al.(2025)Black, Brown, Driess, et~al.]{black2025pi0}
Kevin Black, Noah Brown, Danny Driess, et~al.
\newblock $\pi_0$: A vision-language-action flow model for general robot
  control.
\newblock In \emph{RSS}, 2025.
\newblock arXiv:2410.24164.

\bibitem[Bogenberger et~al.(2026)Bogenberger, Harrison, Dahanaggamaarachchi,
  Brunke, Qian, Zhou, and Schoellig]{bogenberger2026glasses}
Benjamin Bogenberger, Oliver Harrison, Orrin Dahanaggamaarachchi, Lukas Brunke,
  Jingxing Qian, Siqi Zhou, and Angela~P. Schoellig.
\newblock Where did {I} leave my glasses? {O}pen-vocabulary semantic
  exploration in real-world semi-static environments.
\newblock \emph{IEEE Robotics and Automation Letters}, 11\penalty0
  (3):\penalty0 3342--3349, 2026.

\bibitem[Canh et~al.(2026)Canh, Zhang, and Chong]{canh2026hiper}
Thanh~Nguyen Canh, Haolan Zhang, and Nak~Young Chong.
\newblock Which objects stay? {L}earning hierarchical persistence beliefs for
  lifelong open-vocabulary object maps.
\newblock \emph{arXiv:2611.07662}, 2026.

\bibitem[Chow(1970)]{chow1970optimum}
C.~K. Chow.
\newblock On optimum recognition error and reject tradeoff.
\newblock \emph{IEEE Transactions on Information Theory}, 16\penalty0
  (1):\penalty0 41--46, 1970.

\bibitem[Collins et~al.(2022)Collins, Goel, Deng, Luthra, Xu, Gundogdu, Zhang,
  Yago~Vicente, Dideriksen, Arora, Guillaumin, and Malik]{collins2022abo}
Jasmine Collins, Shubham Goel, Kenan Deng, Achleshwar Luthra, Leon Xu, Erhan
  Gundogdu, Xi Zhang, Tomas~F. Yago~Vicente, Thomas Dideriksen, Himanshu Arora,
  Matthieu Guillaumin, and Jitendra Malik.
\newblock {ABO}: Dataset and benchmarks for real-world {3D} object
  understanding.
\newblock In \emph{CVPR}, 2022.

\bibitem[Engelbracht et~al.(2026)Engelbracht, Zurbr{\"u}gg, Wohlrapp,
  B{\"u}chner, Valada, Pollefeys, Blum, and Bauer]{engelbracht2026hoi}
Tim Engelbracht, Ren{\'e} Zurbr{\"u}gg, Matteo Wohlrapp, Martin B{\"u}chner,
  Abhinav Valada, Marc Pollefeys, Hermann Blum, and Zuria Bauer.
\newblock Hoi! -- a multimodal dataset for force-grounded, cross-view
  articulated manipulation.
\newblock In \emph{CVPR}, 2026.
\newblock arXiv:2512.04884.

\bibitem[Ewen et~al.(2024)Ewen, Chen, Chen, Li, Bagali, Gunjal, and
  Vasudevan]{ewen2024feel}
Parker Ewen, Hao Chen, Yuzhen Chen, Anran Li, Anup Bagali, Gitesh Gunjal, and
  Ram Vasudevan.
\newblock You've got to feel it to believe it: Multi-modal {B}ayesian inference
  for semantic and property prediction.
\newblock In \emph{RSS}, 2024.
\newblock arXiv:2402.05872.

\bibitem[Ghosh(2026)]{ghosh2026physvlmm}
Dibyendu Ghosh.
\newblock {PhysVLMM}: Interaction-conditioned physical belief maps for
  articulated objects in robot scene maps.
\newblock \emph{arXiv:2610.40539}, 2026.

\bibitem[Ghosh and Shakya(2026)]{ghosh2026vlmm}
Dibyendu Ghosh and Ayushi Shakya.
\newblock Vision-language-motion maps: An open-vocabulary, uncertainty-aware,
  queryable motion attribute for {3D} scene maps.
\newblock \emph{arXiv:2607.16173}, 2026.

\bibitem[Gu et~al.(2024)Gu, Kuwajerwala, Morin, Jatavallabhula, Sen, Agarwal,
  Rivera, Paul, Ellis, Chellappa, et~al.]{gu2024conceptgraphs}
Qiao Gu, Ali Kuwajerwala, Sacha Morin, Krishna~Murthy Jatavallabhula, Bipasha
  Sen, Aditya Agarwal, Corban Rivera, William Paul, Kirsty Ellis, Rama
  Chellappa, et~al.
\newblock {ConceptGraphs}: Open-vocabulary {3D} scene graphs for perception and
  planning.
\newblock In \emph{ICRA}, 2024.

\bibitem[Hausman et~al.(2015)Hausman, Niekum, Osentoski, and
  Sukhatme]{hausman2015active}
Karol Hausman, Scott Niekum, Sarah Osentoski, and Gaurav~S. Sukhatme.
\newblock Active articulation model estimation through interactive perception.
\newblock In \emph{ICRA}, 2015.

\bibitem[He et~al.(2017)He, Gkioxari, Doll{\'a}r, and Girshick]{he2017maskrcnn}
Kaiming He, Georgia Gkioxari, Piotr Doll{\'a}r, and Ross Girshick.
\newblock Mask {R-CNN}.
\newblock In \emph{ICCV}, pages 2961--2969, 2017.

\bibitem[Kim et~al.(2024)Kim, Pertsch, Karamcheti, Xiao, Balakrishna, Nair,
  Rafailov, Foster, Lam, Sanketi, et~al.]{kim2024openvla}
Moo~Jin Kim, Karl Pertsch, Siddharth Karamcheti, Ted Xiao, Ashwin Balakrishna,
  Suraj Nair, Rafael Rafailov, Ethan Foster, Grace Lam, Pannag Sanketi, et~al.
\newblock {OpenVLA}: An open-source vision-language-action model.
\newblock In \emph{CoRL}, 2024.

\bibitem[Krajn{\'i}k et~al.(2017)Krajn{\'i}k, Fentanes, Santos, and
  Duckett]{krajnik2017fremen}
Tom{\'a}{\v{s}} Krajn{\'i}k, Jaime~Pulido Fentanes, Jo{\~a}o~M. Santos, and Tom
  Duckett.
\newblock {FreMEn}: Frequency map enhancement for long-term mobile robot
  autonomy in changing environments.
\newblock \emph{IEEE Transactions on Robotics}, 33\penalty0 (4):\penalty0
  964--977, 2017.

\bibitem[Li et~al.(2026)Li, Xu, Tian, Sun, Hu, Ye, and Du]{li2026tanav}
Xianyao Li, Fang Xu, Ruitong Tian, Bowen Sun, Xiao Hu, Yang Ye, and Jing Du.
\newblock Nudge before you push: Physics-aware navigation via tactile probing.
\newblock \emph{arXiv:2610.04924}, 2026.

\bibitem[Liu et~al.(2026)Liu, Luo, Liu, Jia, Han, Li, Li, Yang, Zhang, Zheng,
  Xu, and Chong]{liu2026aquamend}
Yufan Liu, Shang Luo, Yang Liu, Haoxuan Jia, Feiyu Han, Qian Li, Chen Li,
  Yingguang Yang, Chongyang Zhang, Hao Zheng, Kefu Xu, and Bin Chong.
\newblock {AquaMend}: Minimal re-probing and conditional rollback for
  latent-belief failures in embodied agents.
\newblock \emph{arXiv:2609.28973}, 2026.

\bibitem[Lo(1987)]{lo1987semiparametric}
Andrew~W. Lo.
\newblock Semi-parametric upper bounds for option prices and expected payoffs.
\newblock \emph{Journal of Financial Economics}, 19\penalty0 (2):\penalty0
  373--387, 1987.

\bibitem[Long et~al.(2021)Long, Rauch, Zhang, Ivan, and
  Vijayakumar]{long2021rigidfusion}
Ran Long, Christian Rauch, Tianwei Zhang, Vladimir Ivan, and Sethu Vijayakumar.
\newblock {RigidFusion}: Robot localisation and mapping in environments with
  large dynamic rigid objects.
\newblock \emph{IEEE Robotics and Automation Letters}, 6\penalty0 (2):\penalty0
  3703--3710, 2021.

\bibitem[Mabrok(2026)]{mabrok2026l2a}
Mohamed~A. Mabrok.
\newblock Open-world law-to-action: Vision--language priors and touch for parts
  with hidden physics.
\newblock \emph{arXiv:2610.33705}, 2026.

\bibitem[Mart{\'i}n-Mart{\'i}n and Brock(2014)]{martinmartin2014online}
Roberto Mart{\'i}n-Mart{\'i}n and Oliver Brock.
\newblock Online interactive perception of articulated objects with multi-level
  recursive estimation based on task-specific priors.
\newblock In \emph{IROS}, 2014.

\bibitem[Milgrom(1981)]{milgrom1981good}
Paul~R. Milgrom.
\newblock Good news and bad news: Representation theorems and applications.
\newblock \emph{The Bell Journal of Economics}, 12\penalty0 (2):\penalty0
  380--391, 1981.

\bibitem[Mur-Artal and Tard{\'o}s(2017)]{murartal2017orbslam2}
Ra{\'u}l Mur-Artal and Juan~D. Tard{\'o}s.
\newblock {ORB-SLAM2}: An open-source {SLAM} system for monocular, stereo, and
  {RGB-D} cameras.
\newblock \emph{IEEE Transactions on Robotics}, 33\penalty0 (5):\penalty0
  1255--1262, 2017.

\bibitem[Nasiriany et~al.(2024)Nasiriany, Maddukuri, Zhang, Parikh, Lo, Joshi,
  Mandlekar, and Zhu]{nasiriany2024robocasa}
Soroush Nasiriany, Abhiram Maddukuri, Lance Zhang, Adeet Parikh, Aaron Lo,
  Abhishek Joshi, Ajay Mandlekar, and Yuke Zhu.
\newblock {RoboCasa}: Large-scale simulation of household tasks for generalist
  robots.
\newblock In \emph{RSS}, 2024.

\bibitem[{Octo Model Team}(2024)]{octo2024}
{Octo Model Team}.
\newblock Octo: An open-source generalist robot policy.
\newblock In \emph{RSS}, 2024.

\bibitem[Perrett et~al.(2025)Perrett, Darkhalil, Sinha,
  et~al.]{perrett2025hdepic}
Toby Perrett, Ahmad Darkhalil, Saptarshi Sinha, et~al.
\newblock {HD-EPIC}: A highly-detailed egocentric video dataset.
\newblock In \emph{CVPR}, 2025.

\bibitem[Qian et~al.(2026)Qian, Du, Yang, Zhou, Waslander, and
  Schoellig]{qian2026active}
Jingxing Qian, Xintong Du, Jun Yang, Siqi Zhou, Steven~L. Waslander, and
  Angela~P. Schoellig.
\newblock Object- and task-aware active perception for safe navigation of
  mobile manipulators in semi-static environments.
\newblock \emph{arXiv:2610.29964}, 2026.

\bibitem[Rosen et~al.(2016)Rosen, Mason, and Leonard]{rosen2016persistence}
David~M. Rosen, Julian Mason, and John~J. Leonard.
\newblock Towards lifelong feature-based mapping in semi-static environments.
\newblock In \emph{ICRA}, pages 1063--1070, 2016.

\bibitem[Saavedra-Ruiz et~al.(2025)Saavedra-Ruiz, Nashed, Gauthier, and
  Paull]{saavedra2025perpetua}
Miguel Saavedra-Ruiz, Samer~B. Nashed, Charlie Gauthier, and Liam Paull.
\newblock {Perpetua}: Multi-hypothesis persistence modeling for semi-static
  environments.
\newblock In \emph{IROS}, 2025.
\newblock arXiv:2507.18808.

\bibitem[Saavedra-Ruiz et~al.(2026)Saavedra-Ruiz, Gauthier, Gupta, Shahfar,
  Ellis, Parkison, and Paull]{saavedra2026predictive}
Miguel Saavedra-Ruiz, Charlie Gauthier, Kumaraditya Gupta, Shima Shahfar,
  Kirsty Ellis, Steven Parkison, and Liam Paull.
\newblock Predictive spatio-temporal scene graphs for semi-static scenes.
\newblock \emph{IEEE Robotics and Automation Letters}, 11\penalty0
  (12):\penalty0 14524--14531, 2026.

\bibitem[Scarf(1958)]{scarf1958minmax}
Herbert Scarf.
\newblock A min-max solution of an inventory problem.
\newblock In \emph{Studies in the Mathematical Theory of Inventory and
  Production}, pages 201--209. Stanford University Press, 1958.

\bibitem[Shukor et~al.(2025)Shukor, Aubakirova, Capuano, Kooijmans, Palma,
  Zouitine, Aractingi, Pascal, Russi, Marafioti, Alibert, Cord, Wolf, and
  Cadene]{shukor2025smolvla}
Mustafa Shukor, Dana Aubakirova, Francesco Capuano, Pepijn Kooijmans, Steven
  Palma, Adil Zouitine, Michel Aractingi, Caroline Pascal, Martino Russi,
  Andres Marafioti, Simon Alibert, Matthieu Cord, Thomas Wolf, and Remi Cadene.
\newblock {SmolVLA}: A vision-language-action model for affordable and
  efficient robotics.
\newblock \emph{arXiv:2506.01844}, 2025.

\bibitem[Singh et~al.(2026)Singh, Happi, Relan, Nair, and
  Baradwaj]{singh2026bokbo}
Anya Singh, Cabrel Happi, Jai Relan, Varun Nair, and Vidyut Baradwaj.
\newblock {BOKBO} (best of {K} bad options): Calibrated abstention for {VLA}
  policies.
\newblock \emph{arXiv:2605.30660}, 2026.

\bibitem[Sturm et~al.(2011)Sturm, Stachniss, and
  Burgard]{sturm2011probabilistic}
J{\"u}rgen Sturm, Cyrill Stachniss, and Wolfram Burgard.
\newblock A probabilistic framework for learning kinematic models of
  articulated objects.
\newblock \emph{Journal of Artificial Intelligence Research}, 41:\penalty0
  477--526, 2011.

\bibitem[Tao et~al.(2025)Tao, Xiang, Shukla, Qin, Hinrichsen, Yuan, Bao, Lin,
  Liu, Chan, et~al.]{tao2025maniskill3}
Stone Tao, Fanbo Xiang, Arth Shukla, Yuzhe Qin, Xander Hinrichsen, Xiaodi Yuan,
  Chen Bao, Xinsong Lin, Yulin Liu, Tse-Kai Chan, et~al.
\newblock Demonstrating {GPU} parallelized robot simulation and rendering for
  generalizable embodied {AI} with {ManiSkill3}.
\newblock In \emph{RSS}, 2025.

\bibitem[Toris and Chernova(2017)]{toris2017temporal}
Russell Toris and Sonia Chernova.
\newblock Temporal persistence modeling for object search.
\newblock In \emph{ICRA}, 2017.

\bibitem[Wang et~al.(2026)Wang, Tian, Swann, Shorinwa, Wu, and
  Schwager]{wang2026phys2real}
Maggie Wang, Stephen Tian, Aiden Swann, Ola Shorinwa, Jiajun Wu, and Mac
  Schwager.
\newblock {Phys2Real}: Fusing {VLM} priors with interactive online adaptation
  for uncertainty-aware sim-to-real manipulation.
\newblock In \emph{ICRA}, 2026.
\newblock arXiv:2510.11689.

\bibitem[Yuan and Holladay(2026)]{yuan2026wontknow}
Linfei Yuan and Rachel Holladay.
\newblock I won't know until {I} try: Interactive information gathering for
  long-horizon manipulation.
\newblock \emph{arXiv:2610.33959}, 2026.

\bibitem[Zhang(2026)]{zhang2026rcdh}
Heng Zhang.
\newblock Do not cut when uncertain: Rejectable and calibrated decision heads
  for {VLA} policies in robotic harvesting.
\newblock \emph{arXiv:2609.35039}, 2026.

\bibitem[Zhou and Li(2026)]{zhou2026trading}
Cheng Zhou and Jing Li.
\newblock Trading transparency for stability: Friction-cone-constrained
  bilateral teleoperation for opening doors of unsecured furniture.
\newblock \emph{arXiv:2611.00898}, 2026.

\bibitem[Zhou et~al.(2025)Zhou, Xiong, Peng, Tao, Zhao, Dong, Tang, and
  Wang]{zhou2025physvlmavr}
Weijie Zhou, Xuantang Xiong, Yi Peng, Manli Tao, Chaoyang Zhao, Honghui Dong,
  Ming Tang, and Jinqiao Wang.
\newblock {PhysVLM-AVR}: Active visual reasoning for multimodal large language
  models in physical environments.
\newblock In \emph{NeurIPS}, 2025.
\newblock arXiv:2510.21111.

\end{thebibliography}
}

\twocolumn[\centering\Large\textbf{Appendix}\par\vspace{1.0em}]  %
\appendix

\section{Method details}
\label{sec:s-method}

\paragraph{Opening controller.} Opening follows a force-limited Cartesian impedance law at the handle point $\mathbf{e}_t$,
\begin{equation}
\mathbf{F}_t=\operatorname{sat}_{F_{\max}}\!\big(k_p(\mathbf{e}^{\star}_t-\mathbf{e}_t)+k_d(\dot{\mathbf{e}}^{\star}_t-\dot{\mathbf{e}}_t)\big),
\label{eq:action}
\end{equation}
where $\operatorname{sat}_{F}$ clips the norm to $F$. The target $\mathbf{e}^{\star}_t$ moves along $\mathbf{d}_i(q)$ at a
speed that ramps up to a constant over $1$\,s, and its lead $\mathbf{e}^{\star}_t-\mathbf{e}_t$ is clipped to $30\,\mathrm{N}/k_p$. Nudge and brace are commanded directly as constant forces.

\paragraph{Proof of \cref{prop:opposite}.} The likelihood ratio of \emph{moved} to \emph{stayed} is non-increasing in $x^{o}$, so the posterior of
$x^{o}$ after \emph{moved} is stochastically smaller than after \emph{stayed}~\cite{milgrom1981good}, and the prior is
their mixture. $\Pb(x^{o}\le\log F_{\max})$ is a decreasing functional and $\chi$ is independent of $x^{o}$, which gives
the first chain. $z$ depends on $\mathbf{x}$ only through $x^{o}$, whose marginal is the same in every component of
\eqref{eq:mix}, so the component weights do not change. In each component with $\varrho_k\ge0$, $x^{c}\mid x^{o}$ is
stochastically increasing in $x^{o}$, so $\Pb(x^{c}>\log F_L)$ follows the order of $x^{o}$. Averaging over components
keeps the order. After the opening update, every posterior is multiplied by the same indicator of $x^{o}\le\log F_{\max}$,
which keeps the likelihood-ratio order of $x^{o}$ and leaves $x^{c}\mid x^{o}$ unchanged, so the same argument
applies.\hfill$\square$

\paragraph{Proof of \cref{prop:bound}.} The value with $Z$ is $(r_+-r_-)\E[(p-p^{\star})^+]$ and without it $(r_+-r_-)(p_0-p^{\star})^+$. With
$\xi=p-p^{\star}$ and $\xi^+-\xi=(-\xi)^+$, both cases reduce to the expected positive part of a variable with variance $\sigma_p^2$ and mean $\kappa$ below zero, which is at most $(\sqrt{\sigma_p^2+\kappa^2}-\kappa)/2$~\cite{scarf1958minmax,lo1987semiparametric}. For the Gaussian link,
$\E[p^2]$ is the probability that two conditionally independent copies of $\tilde{x}^{c}$ given $\tilde{x}^{o}$ both exceed $\theta$; the copies
have correlation $\varrho^2$. As $\partial_r\Phi_2(-\theta,-\theta;r)=e^{-\theta^2/(1+r)}/(2\pi\sqrt{1-r^2})\le1/(2\pi\sqrt{1-r^2})$, integrating over
$r\in[0,\varrho^2]$ gives the $\arcsin$ bound. An observation that is a function of $\tilde{x}^{o}$ reveals less than $\tilde{x}^{o}$ itself, so by the law of total variance its $\sigma_p^2$ is
no larger.\hfill$\square$

\paragraph{Proof of \cref{prop:degenerate}.} $\bar{\mathbf{u}}^{\top}[\mathbf{t}]_{\times}\bar{\mathbf{u}}=\bar{\mathbf{u}}\cdot(\mathbf{t}\times\bar{\mathbf{u}})=0$
for all $\bar{\mathbf{u}},\mathbf{t}$. So a hypothesis with $\mathbf{R}=\mathbf{I}$ counts every such feature in its consensus (inlier) set, which is then
larger than that of the true motion; its rotation error is the angle of $\mathbf{R}_t$.\hfill$\square$

\paragraph{Tracking model.} Each tracked feature, with flow $\Delta\mathbf{u}_j=\mathbf{u}'_j-\mathbf{u}_j$, has a
hidden label $\zeta_j\in\{co,w,out\}$ (moves with the camera, static world, outlier) and the likelihood
\begin{equation}
\begin{split}
p(\mathbf{u}'_j\mid\mathbf{u}_j)=\;&\pi_{co,j}\,\N(\Delta\mathbf{u}_j;\mathbf{0},\sigma_{co}^2\mathbf{I})\\
&+\pi_{w,j}\,\frac{\N(d_E(\mathbf{u}_j,\mathbf{u}'_j);0,s_w^2)}{D_{\mathcal{C}}}+\frac{\pi_{out}}{|\mathcal{C}|},
\end{split}
\label{eq:mixture}
\end{equation}
with $d_E$ the (unsigned) Sampson distance to $E$ in pixels and $\pi_{w,j}=\max(1-\pi_{co,j}-\pi_{out},10^{-6})$. The
wrist camera's valid pixels lie in a circle $\mathcal{C}$ of radius $0.47$ image widths; $D_{\mathcal{C}}$ is its diameter and $|\mathcal{C}|$ its area.

With $\alpha=(\text{grasp state},m_i)$, the prior that a feature in image cell $b$ moves with the camera is the Beta posterior mean
\begin{equation}
\pi_{co}(b\mid\alpha)=\frac{C_{\alpha}(b)+\gamma\,\bar\pi_{\alpha}}{N_{\alpha}(b)+\gamma},
\label{eq:coprior}
\end{equation}
from the $C_\alpha(b)$ of $N_\alpha(b)$ training features of that cell labelled as co-moving, shrunk with strength
$\gamma$ to the action's overall rate $\bar\pi_\alpha$. Before $E$ is known, the world likelihood near zero flow is the
density $\phi_0$ of world flow at zero, and Bayes' rule on $\zeta_j$ predicts the co-moving features from the action alone,
\begin{equation}
\hat\pi_{co,j}=\frac{\pi_{co,j}\,\N(\Delta\mathbf{u}_j;\mathbf{0},\sigma_{co}^2\mathbf{I})}{\pi_{co,j}\,\N(\Delta\mathbf{u}_j;\mathbf{0},\sigma_{co}^2\mathbf{I})+(1-\pi_{co,j})\,\phi_0},
\label{eq:step0}
\end{equation}
with $\pi_{co,j}=\pi_{co}(b(\mathbf{u}_j)\mid\alpha_t)$. $E$ is initialised by RANSAC on $\{j:\hat\pi_{co,j}<0.5\}$ (all features if fewer than eight), which keeps
the zero-flow features of \cref{prop:degenerate} out of the estimate, and is refined by an EM-like alternation on the three-class model \eqref{eq:mixture}
(posterior of $\zeta_j$ under the current $E$; RANSAC on the features with world posterior above $0.5$; the weights and
variances stay fixed).

\section{Parameter estimation}
\label{sec:s-params}

\paragraph{Force prior \eqref{eq:p4}.}
Per direction $d$ and joint type $m$, $\mu_m$ is the training mean of $x^{d}$, and $(\tau_d^2,\sigma_d^2)$ maximise the
marginal likelihood of the training locations' look-alike groups on a $14\times14$ log grid
($\tau^2\in[0.005,1]$, $\sigma^2\in[0.005,1.5]$), as in PhysVLMM~\cite{ghosh2026physvlmm}. Look-alike groups
are the map's geometric groups (same type, parallel fronts, close together), so a part can be grouped before it has
moved; a test part's predictive \eqref{eq:pred} uses only the measured members of its own group in its own location.

\paragraph{Link samples \eqref{eq:mix}.}
For a held-out location, every training location is itself held out in turn (nested) to obtain standardised residuals
$\tilde{x}^{o},\tilde{x}^{c}$ of its parts under a model that excludes both locations. Locations are resampled with replacement $K=200$
times and the residual correlation of each resample is one sample $\varrho_k$, clipped to $[-0.99,0.99]$. The mixture
\eqref{eq:mix} is evaluated on a $281\times281$ grid over $[-2,5]$ (log newtons); prior mass outside
the grid is at most $2.8\,\%$ for one part ($0.7\,\%$ for the class-average prior). On the grid each component is
renormalised, so the marginals agree up to truncation (component weights change by at most $0.2\,\%$). A touch update
\eqref{eq:touch} weights a cell that the threshold cuts by the fraction of the cell on the event's side (mass uniform
within a cell), which avoids the half-cell bias of a mask at cell centres. The perfect-link
ablation ($\varrho=1$, a singular Gaussian) places $x^{c}$ on the line $\hat{x}^{c}+\sqrt{v^{c}/v^{o}}\,(x^{o}-\hat{x}^{o})$
by linear interpolation, with the mass beyond the grid edge (at most $0.04\,\%$) placed on the edge.

\paragraph{Still-in-place model \eqref{eq:renewal}.}
Given $\psi$, the rest gaps are independent and $\Pb(R>a+\Delta\mid R>a,\psi)=S(a+\Delta\mid\psi)/S(a\mid\psi)$; Bayes'
rule over $\psi$ with the likelihood $\prod_j f(r_{i,j}\mid\psi)\,S(a_i\mid\psi)$ gives \eqref{eq:presence}.
The hyper-parameters $(\nu_h,\eta,\omega,w_f,\lambda_f)$ maximise, by L-BFGS, the marginal likelihood of the training kitchens'
completed gaps and of each object's final, right-censored rest period. The label means $\mu_\ell$ and the object rates $\psi_i$ are integrated out on a grid of $151$ values of
$\psi\in[-14,1]$ (log s$^{-1}$). A weak prior $\nu_h\sim\N(\nu_0,3^2)$ on each category mean ($\nu_0$ the
log of the training event rate) keeps single-class categories bounded; $\eta$ and $\omega$ are floored at $0.05$. No method uses the time before an object's first move: objects are annotated only once moved, so that interval would be biased by selection.

\paragraph{Tracking parameters.}
They are fitted on the labelled training sessions, with the test session held out. $\sigma_{co}^2$ is the maximum-likelihood variance of the
co-moving flows. $s_w$ is the root mean square Sampson distance (below $10$\,px) of static features to the essential matrix
estimated from static features, floored at $0.3$\,px. $\pi_{out}$ is the share of features that are neither static nor co-moving, and $\phi_0$
the density of static-feature flow within $5$\,px of zero (the co-moving threshold). $C_\alpha(b)$ and $N_\alpha(b)$ are the per-cell counts for
each action state $\alpha\in\{\text{free},\ \text{grasp}\times\{\text{revolute},\text{prismatic},\text{unknown}\}\}$, with
$44$\,px cells and $\gamma=20$; for a grasp with unknown part type the pooled grasp counts are used. Two rounds of expectation-maximisation are used.

\section{The zero-flow degeneracy (\cref{prop:degenerate})}
\label{sec:s-degeneracy}

A synthetic camera like the wrist camera ($f=609.9$\,px, same image size), world points at $2$--$6$\,m, a random $3^\circ$
rotation and a translation with $2$\,cm standard deviation per axis, $0.5$\,px noise, $400$ features of which a share has zero flow (plus noise), as the
held part's do; essential-matrix RANSAC ($1$\,px, $p=0.999$) and pose recovery, $200$ trials per share (\cref{tab:degeneracy}).

\begin{table}[h]
\centering\small
\setlength{\tabcolsep}{3pt}
\caption{\textbf{Zero-flow degeneracy} in a synthetic two-view test ($200$ trials per share of co-moving features). Err.\ $\approx3^\circ$: error within $0.2^\circ$ of the true rotation, \ie locked to
$\mathbf{R}=\mathbf{I}$; flips: error above $90^\circ$.}
\label{tab:degeneracy}
\begin{tabular}{lcccc}
\toprule
Share & Median err.\ ($^\circ$) & Fail ($>2^\circ$) & Err.\ $\approx3^\circ$ & Flips \\
\midrule
0.0 & 0.23 & 16\,\% & 0 & 33 \\
0.2 & 0.19 & 2\,\% & 0 & 5 \\
0.3 & 0.19 & 3\,\% & 1 & 5 \\
0.4 & 0.36 & 38\,\% & 69 & 1 \\
0.45 & 2.95 & 77\,\% & 148 & 0 \\
0.5 & 2.99 & 100\,\% & 194 & 0 \\
0.6 & 3.00 & 100\,\% & 191 & 0 \\
\bottomrule
\end{tabular}
\end{table}

Two failure modes must be separated. At low shares the failures are $180^\circ$ sign flips of the recovered pose (a
small-translation artefact of pose recovery, $33$ of $200$ trials at share $0$), which vanish as the share grows. From a
share of about $0.4$ the estimate locks to $\mathbf{R}=\mathbf{I}$: its error equals the true rotation angle, which is
the mode of \cref{prop:degenerate}. At a share of $0.5$ the zero-flow features ($200$) already match the median number
of inliers of the true motion ($196$), and $194$ of the $200$ trials lock. The proposition's condition is sufficient; in practice
locking begins before it is met.

\section{Move rate and mass}
\label{sec:s-datacheck}

Whether objects that people move often are also light would link the still-in-place belief to the force belief. On
HD-EPIC~\cite{perrett2025hdepic}, $35$ object types were defined by matching the annotated names. For each type, the move
rate per object and hour of video was computed from the $19{,}812$ annotated moves (counted before overlapping tracks are merged). Twenty of the types have a mass from at least three
matching listings of the Amazon Berkeley Objects dataset~\cite{collins2022abo} (median of the listings). Exclusion
rules for listings were set by reading their titles, never move rates, and masses were frozen before any correlation was computed.
Over the $20$ types, the Spearman correlation between move rate and mass is $+0.03$, with $95\,\%$ interval $[-0.16,+0.43]$ from a
bootstrap over kitchens; all four sensitivity variants (zero-truncated Poisson rate, per-kitchen-hour rate, without chairs and bins, hand-held
objects only) include $0$. The interval excludes a negative correlation stronger than $-0.16$, the direction in which often-moved objects would be lighter.
The check is still weak: the $20$ types cover only $15\,\%$ of the moves, because the most-moved kitchenware (bowls, plates, spatulas,
sponges) has no usable mass.

\section{Checks of the propositions}
\label{sec:s-props}

\paragraph{\cref{prop:opposite} on \mba's actual belief.}
The brace ordering requires every link component $\varrho_k\ge0$; the fitted samples contain a few slightly negative
values (bootstrap noise: $61$ of the $3{,}400$ samples, smallest $-0.174$). Both orderings were therefore verified on
\mba's actual belief for every test part in the main setting and in the $105$ force settings of the sweep (Appendix~\ref{sec:s-details}; the sweep's four nudge
costs do not change the belief), $12{,}826$ cases. Both orderings hold before the opening update, and the brace ordering also holds after it, on the belief the brace decision reads: $0$ violations.

\paragraph{\cref{prop:bound}.}
Never violated on $20{,}000$ random posterior distributions and tight (a two-point distribution attains it); the Gaussian
variance formula agrees with Monte Carlo to within $4.3\times10^{-4}$.

\section{Experimental details}
\label{sec:s-details}

\paragraph{Real pulls.}
The handle force is the generalised force along the joint~\cite{ghosh2026physvlmm}: joint frame fitted on the pull itself, sensor-to-gripper
rotation by Procrustes on the pulls of the other parts of the same session, revolute torque divided by the gripper's distance from the
axis. As in PhysVLMM, a pull needs at least two such pulls in its session ($2$ closing pulls excluded). Force needed is
the peak of the $0.1$\,s median-filtered handle force from the start of the pull until the smoothed gripper has moved half
its largest displacement; values at or below $0.2$\,N are set to $0.2$\,N. No robot nudges: a nudge of force $f_n$ ``moved'' the part iff
the part had moved $1$\,cm before the smoothed handle force of the recorded pull reached $f_n$ (the real nudge outcome of \cref{tab:touch}).

Twenty scan instances whose index points to the joint annotation of a part of another class ($12$, excluded as in PhysVLMM) or to none ($8$) are excluded for every method.
This removes $11$ test parts, $8$ of them in the one location (oven 1) that is then left without test parts. The exclusion was added after the main results.
With these instances ($132$ parts in $18$ locations, main setting), \mba scores $0.293$ and never nudging $0.288$ (difference $+0.005$ {\scriptsize$[-0.020,+0.039]$}),
and \mba's lead over the risk query at $\delta=0.1$ is $+0.134$ {\scriptsize$[-0.043,+0.355]$}, not significant.

In the main setting the limits are
$11.9$--$14.0$\,N and the loads $6.0$--$7.2$\,N over the folds, with $f_n=F_{\max}/2$; the fixed setting has
$F_{\max}=8$\,N, $F_L=5$\,N and $f_n=3$\,N. The sweep crosses $F_{\max}\in\{5,8,12,16,20,25,30\}$\,N, $F_L\in\{3,5,7,10,14\}$\,N,
$f_n\in\{0.25,0.5,0.75\}\,F_{\max}$ and $c_n\in\{0.05,0.1,0.2,0.4\}$: $105$ force settings and $420$ settings in all. The main setting and the sweep were added beside the fixed $8$\,N before
any decision outcome was seen. PhysVLMM's risk query answers ``yes'' if
$\Pb(F^{o}\le F_{\max})\ge1-\delta$, ``no'' if it is at most $\delta$, and ``insufficient evidence'' otherwise; the robot pulls only on ``yes''.

\paragraph{Wrist camera.}
The images are rectified to a pinhole of $1408^2$ pixels. Every third frame of the first $8$ interaction windows of the session is used. A
pair is grasping if the gripper is closed in both frames. Pairs with less than $1^\circ$ of true rotation are excluded, which leaves $11{,}190$ pairs, $2{,}702$ of them grasping (\cref{tab:camera,fig:s-exp2} use these). Ground truth comes from the device's metric visual-inertial
trajectory with the calibration's pinhole transform (median epipolar error on free pairs $0.55$\,px) and a $5$\,ms
rolling-shutter correction. A feature is static if its endpoint is consistent with a static point on its ray (exact
without depth), and co-moving if it is not static and its flow is below $5$\,px. All methods share a lens mask and a
self-body mask learned per session without ground truth. ORB-SLAM2 and DynaSLAM run from DynaSLAM's released implementation, with build and run-time compatibility fixes only and a driver that changes only input and output (masks, cached Mask R-CNN output). They run monocular with $2{,}000$ ORB features, Hoi!'s camera intrinsics and otherwise the released settings for indoor sequences, three times per session; the \mba mask
covers the cells with co-moving prior above $0.5$.

\paragraph{Kitchen activity.}
Overlapping tracks of an object are merged ($192$ tracks absorbed) and track ends clipped to the video end. At $t=60j$\,s
an object is queried if it is at rest and $t+\Delta$ does not pass the video end. Only tracks started by $t$ are
used. The published methods' settings are chosen per fold by log-loss on the training kitchens.
Persistence filter: one filter per rest period, started at the placement, with a detection at every grid time up to
$t$. Its parameters $(\lambda_l,\lambda_u,P_M,P_F)$ come from a grid of $48$; every fold chooses $(10^{-5},1,0.1,0.01)$. Three of these lie on the grid edge, and a wider grid of $144$ leaves the score unchanged.
FreMEn: the binary state ``an object of this class at rest at time $s$ is not
picked up within $\Delta$'', per kitchen and class or per kitchen. At each query time a new model is given all earlier
samples. Order $0$--$4$ and granularity are chosen per fold; every fold chooses order $0$ (the mean only, no periodic term) per kitchen.
Perpetua: its released code, driven like the persistence filter. Its lifetime mixtures are fitted by its own EM on
the training kitchens' rest periods, cut at the annotated placements and pick-ups (one ``still in place'' reading just
before each pick-up), with $1$--$3$ components chosen by AIC as in its released example. The prior family and
$(P_M,P_F)$ are chosen from three of the four combinations of its paper's and examples' defaults (exponential or log-normal; $0.1$ or $0.01$). The fourth, the worst on the training kitchens in a quick check, was dropped to bound the run time. Every fold chooses exponential with $P_M=P_F=0.01$. Its re-appearance part never
becomes active here.

\paragraph{Simulation.}
ManiSkill3 on the CPU at $100$\,Hz, $10$ layouts $\times$ $12$ styles, each part rebuilt with its fitted physics.
Controller \eqref{eq:action}: $k_p=300$\,N/m, $k_d=60$\,N\,s/m, commands at $10$\,Hz, target speed $0.15$\,m/s. Nudge $0.5$\,s; brace after a $0.5$\,s hold; moved and slipped mean at least $1$\,cm; a part's support is the largest constant
load it holds still for $1$\,s. Layouts $0$--$6$ are used for training and layouts $7$--$9$ for testing, one episode per part and method. Main
setting: $F_{\max}=4.62$, $F_L=2.76$, $f_n=2.31$\,N. It uses the real-pull rule and was added,
after an $80$-episode pilot on two test kitchens, because the fixed $8$\,N left nothing to decide; both settings are
reported, and Appendix~\ref{sec:supp-results} gives the main setting without the two kitchens. Methods held to the safe limit cause no damage by construction, since the damage measure counts force above that limit.
The $0.5$\,s in the damage measure is about the time the controller needs to wind from the safe limit to $30$\,N on a part that does not move ($0.5$--$0.7$\,s, depending on the limit and on whether the part was moving).

\mba's belief here is simpler than in \cref{sec:belief}: it links the two forces by one linear fit per part kind on the
training layouts (correlation $0.51$--$0.94$, from the simulator's shared Coulomb friction), with one $\varrho$ per part
and $\beta$ per kind. The blocked state used to fit $\beta$ is the blocked flag of the simulated scenes (a part that
stalls under a $20$\,N ramp~\cite{ghosh2026physvlmm}). Under the $30$\,N controller, $64\,\%$ of the flagged doors and $74\,\%$ of the flagged drawers
open, so flag and outcome disagree: $\beta$ overstates blocking for drawers ($0.08$ vs $0.03$) and understates it for doors ($0.05$ vs $0.09$).
Using the closed-loop rate per kind instead would change $64$ open and $187$ nudge
decisions in the main setting ($1$ and $0$ at $8$\,N). We kept the $\beta$ fixed before the runs; the score under the other rate would need new simulator runs and was not computed.
In the main setting, the planner's opening model, $\chi=0$ and $F^{o}\le F_{\max}$, disagrees with the simulator's
outcome on $118$ of the $1{,}254$ test parts whose measured effort settled; the other $179$ test parts have no such force. The
simulated nudge differs from the threshold event $F^{o}\le f_n$ on $28$ of \mba's $181$ nudges ($3$ of them on parts without a settled force); on $26$ the part moved although
$F^{o}>f_n$.

Demonstrations: the scripted controller at $30$\,N on $1{,}008$ training parts, $987$ successful episodes.
The input is a $224^2$ image from a fixed camera, the handle displacement and ``open the door/drawer''; the action is the target increment
per $0.1$\,s. SmolVLA is fine-tuned from its released base model (batch $16$, $20{,}000$ steps), and Octo (small, version $1.5$)
with its official fine-tuning example (batch $8$, $40{,}000$ steps, the same number of samples). $\pi_0$ and OpenVLA did not
fit in GPU memory beside the simulator, so they were not run.

\paragraph{Tests and run time.}
All experiments ran on a single consumer laptop. The computations were checked on cases with known answers, such as the planner against brute-force search over all
policies. Median times per case on one CPU thread are as follows.
\mba takes $0.74$\,ms for a still-in-place query, $56$\,ms for a part's force belief, $1.1$\,ms for its plan, and $12.4$\,ms per frame pair for tracking (after the
shared feature tracker).
The same query takes $0.57$\,ms for the persistence filter and $3.9$\,ms for FreMEn. When a kitchen's queries are batched, the times are $11$,
$18$ and $73$\,\textmu s for \mba, the persistence filter and FreMEn. Fitting
\mba's still-in-place model takes $8.4$\,s per fold, and the simulation planner $2.9$\,ms per part. Two-view odometry takes $6.1$\,ms per frame
pair, after the shared feature tracker ($48$\,ms). ORB-SLAM2 and DynaSLAM track a frame in $50$ and $67$\,ms
(multi-threaded, within the $0.1$\,s budget of real time), and Mask R-CNN~\cite{he2017maskrcnn}, run once beforehand, adds $3.9$\,s. The policies were not timed.

\section{Additional results}
\label{sec:supp-results}

\begin{figure*}[t]
\centering
\includegraphics[width=\textwidth]{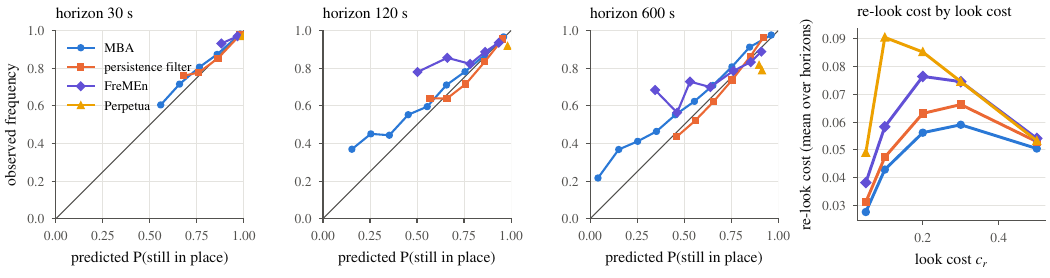}
\vspace{-1.5em}
\caption{\textbf{Kitchen activity: calibration and re-look cost.} Reliability per horizon (bins with at least
$200$ queries) and re-look cost against the look cost $c_r$, mean over horizons. Perpetua's predictions lie in a narrow high range, so it has one or two bins per horizon.}
\label{fig:s-exp3}
\end{figure*}

\begin{figure*}[t]
\centering
\includegraphics[width=\textwidth]{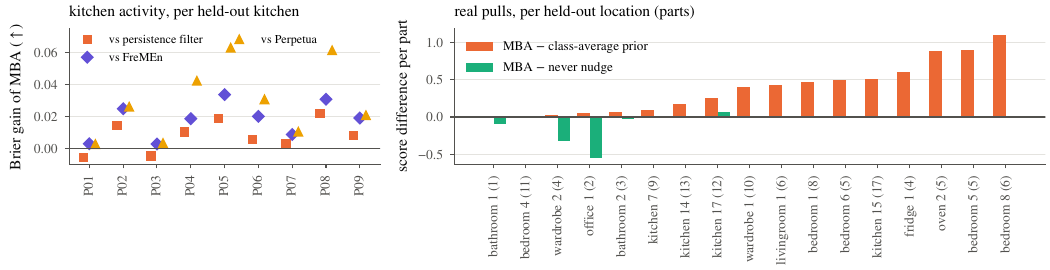}
\vspace{-1.5em}
\caption{\textbf{Per held-out kitchen and location.} Left: Brier gain of \mba over each published method per kitchen (kitchen activity; P01--P09 are the HD-EPIC participants' kitchens). Right: score
difference per part in each location, \mba minus the class-average prior and \mba minus never nudging (real pulls, main setting; parts in brackets).}
\label{fig:cluster}
\end{figure*}

\paragraph{Kitchen activity.} The gap to the published methods grows with the horizon $\Delta$ (\cref{tab:horizon}). Over the kitchens, the margin in re-look cost over the persistence filter has paired
$t$-test $p=0.013$, Wilcoxon $p=0.020$ and sign test $7/9$, $p=0.18$; in Brier score, Wilcoxon $p=0.055$. Against Perpetua, \mba is ahead in all $9$ kitchens (\cref{fig:cluster}; Brier: paired $t$-test $p=0.005$, Wilcoxon and sign test $p=0.004$, the smallest possible with $9$ kitchens). The
persistence filter as run predicts almost like the time-since-placement control (correlation $0.990$). \mba's re-look cost is below that of the three published methods at every look cost $c_r$ (\cref{fig:s-exp3}). Where \mba
gives a low chance of being in place at $120$ and $600$\,s, that chance is too low (at $600$\,s, of objects given $0.04$, $22\,\%$ are still in place).
The persistence filter makes no such low predictions and is close to calibrated where it predicts.

\begin{table}[t]
\centering\small
\setlength{\tabcolsep}{3pt}
\caption{\textbf{Kitchen activity per horizon} ($127{,}499$, $115{,}157$ and $66{,}932$ queries). Brier: mean squared error of the predicted probability; re-look: re-look cost, mean over the look costs $0.05$--$0.5$.}
\label{tab:horizon}
\resizebox{\columnwidth}{!}{%
\begin{tabular}{lcccccc}
\toprule
& \multicolumn{2}{c}{$30$\,s} & \multicolumn{2}{c}{$120$\,s} & \multicolumn{2}{c}{$600$\,s} \\
\cmidrule(lr){2-3}\cmidrule(lr){4-5}\cmidrule(lr){6-7}
Method & Brier & Re-look & Brier & Re-look & Brier & Re-look \\
\midrule
\mba & .0271 & .0199 & .0659 & .0441 & .1357 & .0776 \\
Persistence filter & .0276 & .0207 & .0692 & .0478 & .1466 & .0881 \\
FreMEn & .0293 & .0233 & .0755 & .0561 & .1666 & .1016 \\
Perpetua & .0299 & .0232 & .0800 & .0636 & .1789 & .1247 \\
\bottomrule
\end{tabular}}
\end{table}

\begin{figure}[t]
\centering
\includegraphics[width=\columnwidth]{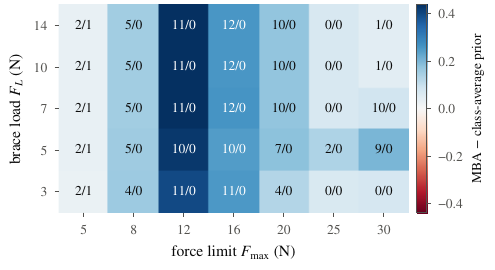}
\vspace{-1.5em}
\caption{\textbf{Real pulls: sweep against the class-average prior.} Colour: mean score difference over the $12$
settings per cell ($3$ nudge forces $\times$ $4$ nudge costs); text: settings with \mba better / worse (interval excluding $0$).}
\label{fig:s-sweep}
\end{figure}

\paragraph{Real pulls.} Against never nudging, \mba ties in $8$ locations, is ahead in $1$ (with $12$ parts) and
behind in $8$, by $0.01$--$0.55$ per part; most of the loss is in two locations, with $2$ and $4$ parts (\cref{fig:cluster}). All sweep losses against the class-average prior are at
$F_{\max}=5$\,N, where almost nothing opens (\cref{fig:s-sweep}). Over the sweep, \mba braces in $7.7\,\%$ of cases ($28\,\%$ at $F_L=3$\,N, none at $10$\,N or more), and $9.5\,\%$ of its braces slip. After the part has opened, and weighted by the chance of opening, a further exact observation of the opening force is worth $0.0026$
per part (\cref{prop:bound} allows at most $0.020$), or $0.0033$ per brace decision (at most $0.031$); it beats the
nudge cost on two parts only. Against PhysVLMM's risk query at $\delta=0.1$, \mba is ahead in $9$ locations, tied in $2$ and
behind in $6$; its gain grows with the force limit. With $\delta=0.2$ and $0.3$ the query scores $0.165$ and $0.207$ (\mba $0.304$).

\paragraph{Costs.} We varied the failed-pull cost $r_f\in\{-0.5,-1,-3,-9\}$ and the slip cost $r_s\in\{-1,-3,-9\}$ in the main setting, with the other costs of \cref{tab:setup}.
\mba is significantly ahead of the risk query at $\delta=0.1$ when a failed pull costs at most $1$ ($+0.22$ to $+0.26$), and ahead but not significantly when it costs $3$ ($+0.04$ to $+0.06$).
At $r_f=-9$ the costs themselves imply $\delta=0.1$, and both score below never pulling (\mba $-0.16$ to $-0.17$, the query $-0.18$): of the $18$ parts to which the map prior gives $\Pb(\text{open})\ge0.9$, $4$ fail (in two locations) where about $1$ is expected.
Against the query at the $\delta$ the costs imply, \mba never differs significantly ($-0.07$ to $+0.03$), and the slip cost moves \mba's score by at most $0.02$.

\begin{table}[!t]
\centering\small
\setlength{\tabcolsep}{3pt}
\caption{\textbf{Wrist camera} ($2{,}702$ pairs while holding a part). Failed: rotation error above $2^\circ$ or no pose; tracked: share of pairs with a pose.
Differences in percentage points, against \mba (top) or ORB-SLAM2 with \mba's mask (bottom); bold: interval
excluding $0$; ref.: uses ground-truth labels, for reference only. SLAM: mean of three runs (range over runs: ORB-SLAM2 $82.5$--$85.0\,\%$, DynaSLAM $83.9$--$84.8\,\%$, with \mba's mask $80.7$--$84.2\,\%$).}
\label{tab:camera}
\resizebox{\columnwidth}{!}{%
\begin{tabular}{lccl}
\toprule
Method & Failed\,$\downarrow$ & Tracked & Method $-$ reference \\
\midrule
\mba & 35.5\,\% & 100\,\% & -- \\
Without the action prior (abl.) & 34.7\,\% & 100\,\% & $-0.8$ {\scriptsize$[-2.3,+0.6]$} \\
No part type (abl.) & 35.5\,\% & 100\,\% & $0.0$ {\scriptsize$[-0.5,+0.6]$} \\
No image-position prior (abl.) & 35.5\,\% & 100\,\% & $0.0$ {\scriptsize$[-1.4,+1.3]$} \\
Two-view odometry (ctrl.) & 38.3\,\% & 100\,\% & $+2.8$ {\scriptsize$[-1.1,+6.3]$} \\
Drop true co-moving (ref.) & 33.9\,\% & 99\,\% & $\mathbf{-1.6}$ {\scriptsize$[-3.0,-0.4]$} \\
Keep only true static (ref.) & 27.2\,\% & 86\,\% & $\mathbf{-8.4}$ {\scriptsize$[-11.3,-4.7]$} \\
\midrule
ORB-SLAM2~\cite{murartal2017orbslam2} & 83.7\,\% & 19\,\% & $+1.1$ {\scriptsize$[-0.6,+2.9]$} \\
DynaSLAM~\cite{bescos2018dynaslam} & 84.3\,\% & 18\,\% & $+1.7$ {\scriptsize$[-0.6,+4.1]$} \\
ORB-SLAM2 + \mba mask & 82.6\,\% & 19\,\% & -- \\
\bottomrule
\end{tabular}}
\end{table}

\begin{figure}[t]
\centering
\includegraphics[width=\columnwidth]{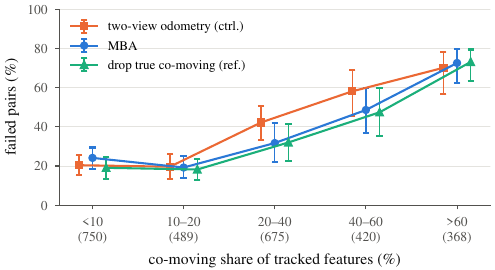}
\vspace{-1.5em}
\caption{\textbf{Wrist camera: two-view failures by the share of co-moving features} (pairs per bin in brackets; $95\,\%$ intervals over locations).}
\label{fig:s-exp2}
\end{figure}

\paragraph{Wrist camera.} Even with ground-truth labels, removing the held part's features lowers failures only
slightly (\cref{tab:camera}). Keeping only the true static features lowers them most, partly because the labels also select pairs whose
ground truth is consistent. The remaining failures appear to come from the part moving relative to
the gripper, tracking errors, blur and label noise (not quantified per cause). \mba and removing the true co-moving features help most when
$20$--$60\,\%$ of the features move with the camera. Above $60\,\%$, too few static features remain and every method fails
in about $70\,\%$ of pairs (\cref{fig:s-exp2}).

\begin{table}[t]
\centering\small
\setlength{\tabcolsep}{2.5pt}
\caption{\textbf{Simulation per part kind} (main setting, safe limit; test parts in brackets).}
\label{tab:kind}
\resizebox{\columnwidth}{!}{%
\begin{tabular}{lcccccc}
\toprule
& \multicolumn{2}{c}{Doors ($893$)} & \multicolumn{2}{c}{Drawers ($504$)} & \multicolumn{2}{c}{Microwaves ($36$)} \\
\cmidrule(lr){2-3}\cmidrule(lr){4-5}\cmidrule(lr){6-7}
Method & Score & Declined & Score & Declined & Score & Declined \\
\midrule
\mba (scripted motion) & 0.174 & 42\,\% & 0.166 & 69\,\% & 0.381 & 28\,\% \\
SmolVLA + \mba & 0.154 & 42\,\% & 0.170 & 69\,\% & 0.381 & 28\,\% \\
Octo + \mba & 0.190 & 42\,\% & 0.170 & 69\,\% & 0.381 & 28\,\% \\
Class-average prior only (abl.) & 0.216 & 63\,\% & 0.000 & 100\,\% & 0.389 & 0\,\% \\
Scripted, safe limit (ctrl.) & 0.046 & 0\,\% & 0.226 & 0\,\% & 0.389 & 0\,\% \\
SmolVLA, safe limit & 0.015 & 0\,\% & 0.234 & 0\,\% & 0.389 & 0\,\% \\
Octo, safe limit & 0.084 & 0\,\% & 0.234 & 0\,\% & 0.389 & 0\,\% \\
\bottomrule
\end{tabular}}
\end{table}

\paragraph{Simulation.} Without the two pilot kitchens ($34$ kitchens, $1{,}353$ parts), \mba's gain over pulling at the safe limit is $+0.062$ {\scriptsize$[+0.016,+0.107]$} with SmolVLA, $+0.041$ {\scriptsize$[-0.003,+0.087]$} with Octo and $+0.058$ {\scriptsize$[+0.012,+0.104]$} with scripted motion (all $36$ kitchens: $+0.064$, $+0.043$, $+0.059$). With both policies, \mba ties the risk query at $\delta=0.5$: $+0.017$ {\scriptsize$[-0.003,+0.037]$} with SmolVLA and $+0.008$ {\scriptsize$[-0.009,+0.028]$} with Octo.
On drawers \mba scores below pulling at the safe limit (\cref{tab:kind}). Only $14\,\%$ of training parts hold the fixed $5$\,N
load (\mba's largest $\Pb(\text{hold})$ there is $0.65$ before opening and $0.24$ after it), and about half hold the main setting's load (their median); \mba
never braced. Forcing the controls to brace makes them lose $1.0$--$2.4$ per part through slips, so we do not compare
under forced bracing. At the fixed $8$\,N, $84\,\%$ of parts open within the limit, \mba declines $1\,\%$, and the scores are $0.682$
(scripted), $0.672$ (SmolVLA) and $0.715$ (Octo), with and without \mba. The class-average prior declines every drawer and
scores above \mba on doors (\cref{tab:kind}), so \mba's small lead over it comes from drawers.

\end{document}